\documentclass[conference]{IEEEtran}
\IEEEoverridecommandlockouts
\usepackage{cite}
\usepackage{amsmath,amssymb,amsfonts,amsthm}
\usepackage{subfigure}
\usepackage{multirow}
\usepackage{makecell}
\usepackage{algorithmic}
\usepackage{graphicx}
\usepackage{textcomp}
\usepackage{xcolor}
\def\BibTeX{{\rm B\kern-.05em{\sc i\kern-.025em b}\kern-.08em
    T\kern-.1667em\lower.7ex\hbox{E}\kern-.125emX}}
\usepackage{algorithm}
\usepackage{algorithmic}
\usepackage{multicol}

\newcommand{\RR}{\mathbb{R}}

\usepackage[capitalize,noabbrev]{cleveref}
\newcommand{\ie}{\textit{i.e.}}
\usepackage{blindtext}
\newcommand{\eg}{\textit{e.g.}}

\newtheorem{assumption}{Assumption}
\newtheorem{definition}{Definition}
\newtheorem{lemma}{Lemma}
\newtheorem{theorem}{Theorem}
\newtheorem{proposition}{Proposition}
\newtheorem{corollary}{Corollary}

\newtheorem{claim}{Claim}
\newcommand{\para}[1]{\noindent {\bf #1}}

\newcommand{\sys}{AIMS}

\begin{document}



\title{Aggregation in the Mirror Space (AIMS): \\Fast, Accurate Distributed Machine Learning in  Military Settings}



\author{\IEEEauthorblockN{Ryan Yang}
\IEEEauthorblockA{\textit{Choate Rosemary Hall}}
\and
\IEEEauthorblockN{ Haizhou Du}
\IEEEauthorblockA{
\textit{Shanghai University of Electric Power}
}
\and
\IEEEauthorblockN{Andre Wibisono}
\IEEEauthorblockA{\textit{Yale University}}

\and
\IEEEauthorblockN{Patrick Baker}
\IEEEauthorblockA{\textit{Royal Air Force}}
}


\maketitle

\begin{abstract}
Distributed machine learning (DML) can be an important capability for modern military to take advantage of data and devices distributed at multiple vantage points to adapt and learn. The existing distributed machine learning frameworks, however, cannot realize the full benefits of DML, because they are all based on the simple linear aggregation framework, but linear aggregation cannot handle the \textit{divergence challenges} arising in military settings: the learning data at different devices can be heterogeneous (\ie, Non-IID data), leading to model divergence, but the ability for devices to communicate is substantially limited (\ie, weak connectivity due to sparse and dynamic communications), reducing the ability for devices to reconcile model divergence. 

In this paper, we introduce a novel DML framework called aggregation in the mirror space (AIMS) that allows a DML system to introduce a general mirror function to map a model into a mirror space to conduct aggregation and gradient descent. Adapting the convexity of the mirror function according to the divergence force, AIMS allows automatic optimization of DML. We conduct both rigorous analysis and extensive experimental evaluations to demonstrate the benefits of AIMS. For example, we prove that AIMS achieves a loss of \(O\left((\frac{m^{r+1}}{T})^{\frac1r}\right)\) after \(T\) network-wide updates, where \(m\) is the number of devices and \(r\) the convexity of the mirror function, with existing linear aggregation frameworks being a special case with \(r=2\). 
During weak connectivity allowing only a relatively small number of complete updates (\ie, \(m \ge T\)), traditional linear aggregation \((r=2)\) has a loss that is \(O(\sqrt{m/T})\) times that of AIMS that optimizes by selecting large \(r\).  
Our experimental evaluations using EMANE (Extendable Mobile Ad-hoc Network Emulator) for military communications settings show similar results: AIMS can improve DML convergence rate by up to 57\% and scale well to more devices with weak connectivity, all with little additional computation overhead compared to traditional linear aggregation.

\end{abstract}

\section{Introduction}
\label{sec:introduction}
Machine Learning (ML) is a crucial technology in the military setting. In the civilian context, it has already found widespread usage in solving a wide range of recognition and predictive tasks. The military is no different: sensors collect massive amounts of data, and efficient analysis of the data using machine learning can assist in decision-making. Machine learning advances have facilitated capabilities such as autonomous reconnaissance and attack with a UAV swarm~\cite{Autonomous2021} and unmanned reconnaissance naval ships ~\cite{8599848}. 

One way to realize the aforementioned use cases is distributed machine learning (DML), which can take advantage of data and devices distributed at multiple vantage points to adapt and learn. 
Given the advantage, multiple DML frameworks have been proposed and analyzed, and they can be classified as three main categories: parameter server (\eg, \cite{smola2010architecture}); all-reduce  (\eg, \cite{allreduce2010}) or ring-reduce (\eg, \cite{lee2020tornadoaggregate}); and gossip (\eg, \cite{hegedHus2019gossip}).
The first two categories can produce highly regular traffic patterns and hence can violate important requirements such as low probability of detection (LPD). In this paper, we focus on the gossip category, in which devices share their local models with neighbors and aggregate received local models. 


All three categories aggregate received models with a linear mean (weighted average). Because of this, they may struggle in military settings due to \textit{divergence forces}. First, learning data at different devices can be heterogeneous (\ie, Non-IID data), and Non-IID data can cause models to naturally diverge. Second, the ability for devices to communicate in military settings can be substantially limited. 
Traditional linear aggregation with weak communication cannot effectively prevent the model divergence, so the models drift away from each other, and the local gradient steps at different local models are no longer as interchangeable.


In this paper, we introduce the \sys{} framework which maps models to a mirror space, similar to mirror descent \cite{bubeckReview}, and does both averaging as well as gradient steps in the mirror space. 
For example, when the mapping is $x^p$, the aggregation step becomes a weighted power-$p$ mean.
The \sys{} framework, depending on the choice of mirror mapping, can decrease the variance among the models held at local devices. \sys{} can adapt the convexity of the mirror function in response to the strength of \textit{divergence forces}, with traditional linear aggregation being used in cases with weak \textit{divergence forces}.
These changes translate into real-world benefits: extensive experiments show that under the \sys{} aggregation framework, the convergence speed of DML models in low-communication environments is accelerated by up to 57\% and the accuracy of the trained model is also higher than that achieved by using the traditional linear aggregation.

The \textbf{main contributions} of this paper are as follows:
\begin{itemize}
\setlength\itemsep{0em}
    \item We study the important issue of \textit{diverging forces} in DML in the case of Non-IID data with sparse/dynamic communication networks. This can help boost the performance of decentralized military applications such as autonomous reconnaissance on UAV swarms and mobile image classification on unmanned ships. 
	\item We propose \sys{}, a novel aggregation framework,  that improves the convergence speed and scalability by aggregating in the mirror space. 
	\item We conduct rigorous analysis, and prove that the loss bound of AIMS to be $O((\frac{m^{r+1}}{T})^{\frac1r})$, where $T$ is the number of iterations, $m$ the number of devices, and $r$ the degree of uniform convexity of the mirror function. Large $r$ gives improvement over traditional linear aggregation $(r=2)$ in weak-communication when $T\leq m$. Our theoretical results also recover state of the art bounds in distributed gradient descent, distributed mirror descent, and single-device mirror descent under a generic uniform convexity assumption.
	\item We conduct extensive experiments to evaluate the performance of our framework. Experimental results show that \sys{} framework accelerates the convergence speed by up to 57\%, and improves scalability in low-communication environments.
\end{itemize}


The rest of this paper is organized as follows. Section \ref{sec:backgroud} describes the problem formulation and motivation. Section \ref{sec:design} describes the \sys{} framework. In Section \ref{sec:convergences}, we discuss properties/observations about \sys{}. Section \ref{sec:evaluationss} provides experimental evaluations and we conclude in Section \ref{sec:conclusion}.


\section{Problem Formulation and Motivation}
\label{sec:backgroud}
\subsection{Problem Formulation}
We consider distributed machine learning with $m$ devices and no central servers. In each time interval, we assume that communication channels exist between certain pairs of local devices, and that the channels are symmetric. These situations are common in practical applications (\eg, autonomous reconnaissance by a UAV swarm).

Each device $i$ has its own dataset $D_i$, from which it constructs its local loss function $f_i$. Then, if the model parameters are represented by a vector $\textbf{w}\in \RR^d$, the devices aim to collaboratively solve the following unconstrained optimization problem
\begin{equation}
\label{eq:problem}
    \min F(\textbf{w}) = \sum_{i=1}^m f_i(\textbf{w}),
\end{equation}
without sharing local data.

\subsection{Motivation}
Distributed machine learning aims to solve the aforementioned problem. Assume each device $i$ computes a local model $\textbf{w}_i$ using its dataset $D_i$, and the $D_i$s are heterogeneous; then the $\textbf{w}_i$ will be different. If the connection between local devices is weak, the nodes cannot exchange models, and therefore they will end up with different models. 


The general idea of \sys{} is to introduce a new aggregation method so that different models at different nodes converge faster with the same number of combination steps. Consider the traditional linear aggregation steps of the form $x_{i,t+1} = \sum_{j=1}^m P(t)_{i,j} x_{j,t}$ where $P(t)$ is a doubly stochastic matrix representing connectivity at time $t$. \sys{} maps model values to a mirror space, averages there, and then maps back. In particular, assuming that the mapping is $x\to x^p$, then \sys{} would give a ``weighted power mean":
\begin{equation}
\label{eq:WPM}
    x_{i,t+1} = \left(\sum_{j=1}^m P(t)_{i,j} x_{j,t}^p\right)^{\frac1p}.
\end{equation}
Such mappings help reduce variance compared to traditional linear averaging. 
To build intuition, consider an example setting where device $1$ has model $w_{1,t}=3$, and device $2$ has model $w_{2,t}=11$. Assume the connectivity matrix is $P(t)=\begin{bmatrix} 0.6 & 0.4\\ 0.4 & 0.6 \end{bmatrix}$, under traditional linear aggregation, this gives a final difference of $w_{2,t+1}-w_{1,t+1}= (0.4\cdot 3 + 0.6\cdot 11) - (0.6\cdot 3 + 0.4\cdot 11) = 7.8-6.2 = 1.6$, 
while using a weighted power mean of degree $p=5$ gives a difference of $w_{2,t+1}-w_{1,t+1}=(0.4\cdot 3^5+0.6\cdot 11^5)^{\frac15}-(0.6\cdot 3^5+0.4\cdot 11^5)^{\frac15}\approx 9.93-9.16 = 0.77$. In this case, changing the value of $p$ has more than halved the final difference. 

The weighted power mean is a special case of the general strategy of averaging in a mirror descent dual space. In general, the \sys{} version of the aggregation step is the following equation:
\begin{equation}
    x_{i,t+1} = h^{-1} \left(\sum_{j=1}^m P(t)_{i,j} h(x_{j,t})\right),
\end{equation}
where $h$ is the mapping function. For example, in the preceding example, $h(x)=x^p$.
The choice of $h$ best suited for a given problem may vary.
\section{The \sys{} Framework}
\label{sec:design}
\subsection{Overview}
With the background and motivation presented in the preceding section, we now give the AIMS framework, which specifies the behavior of each device in Algorithm \ref{alg:new}. The framework solves the problem stated in Equation \ref{eq:problem}. 

The framework assumes that device $i$ maintains its own local model denoted by $\textbf{w}_{i}$; for simplicity of specification, Algorithm 1 uses a round-based specification, in which the model holds by device $i$ after $t$ computations is written as $\textbf{w}_{i,t}$. In real execution, the protocol is asynchronous.
 
A key part of the framework is the generation of the $P(t)_{ij}$ matrices. $P(t)\in \RR^{m\times m}$ is used in iteration $t$ to define how each local device aggregates received models. Each $P(t)$ is defined by
\begin{align}
\label{eq:Pt1}
P(t)_{ij}&= \text{min} \{ e_{i,j}(t), e_{j,i}(t) \},\\
\label{eq:Pt2}
P(t)_{i,i} &= 1- \sum_{j\neq i} P(t)_{i,j},
\end{align}
where $e_{i,j}(t) = \frac{1}{N_i(t)+1}$, if $i,j$ are connected and $e_{i,j}(t)=0$ if $i,j$ are not connected. $P(t)$ generated in this way satisfy Assumption \ref{ass:connectivity} (In Section 4.A). 
Note that the connectivity $e_{i,j}(t)$ is determined by underlying communication systems, which consider both feasibility and security requirements (\eg, LPD).
The framework applies to generic DML and the parameter server and all-reduce frameworks can be seen as special cases (where all $P(t)_{ij}$ values equal $\frac1m$).

With $\textbf{w}_{i,t}$ and  $P(t)_{ij}$ defined, in each iteration $t$, each device $i$ first generates a weighted mean $\textbf{y}_{i,t}$,
\begin{equation}
   \textbf{y}_{i,t} = h_t^{-1} (\sum_{j=1}^m P(t)_{ij} h_t(\textbf{w}_{j,t})),
\end{equation}
where $h_t$ is the mapping function determined for each iteration. Then, each device takes a mirror descent step. This generates the new $\textbf{w}_{i,t+1}$ value with a mirror descent step of the form
\begin{equation}
    h_t(\textbf{w}_{i,t+1}) = h_t(\textbf{y}_{i,t}) - \eta \nabla f_i(w_{i,t}).
\end{equation}

To ease analysis, we will consider the execution of Algorithm \ref{alg:new} while all $h_t(x)$ are equal to $h(x)$.
\begin{algorithm}[!hbtp]
\caption{Instructions for each local device. This is a Distributed Protocol under the \sys{} Framework.}
\label{alg:new}
\begin{algorithmic}[1]
\STATE \textbf{Input: } The datasets $D_i$, the initial model parameters $\mathbf{w}_i(0)$, the learning rate $\eta_i(t)$ for each device $i$, $t=0, \ldots, T$.
\STATE \textbf{Output: } The final model parameters of all devices after $T$ iterations $\mathbf{w}_i(t)$.
\FOR{$t=0$ to $T$}
\STATE Compute $N_i(t)$, the number of devices that node $i$ can connect with.
\STATE Send $h(\textbf{w}_{i,t})$ and $e_{i,j}(t) = \frac{1}{N_i(t)+1}$ to $i$'s neighbors.
\STATE Receive $h(\textbf{w}_{j,t})$ values and compute $P(t)_{ij}$ values according to Equation \ref{eq:Pt1} and \ref{eq:Pt2}.
\STATE Compute local subgradient $d_{i}(t)=\nabla f_i (\textbf{w}_{i,t})$ 
\STATE Compute $\textbf{y}_{i,t} = h^{-1} (\sum_{j=1}^m P(t)_{ij} h(\textbf{w}_{j,t}) )$.
\STATE Compute $\textbf{w}_{i,t+1} =h^{-1}( h(\textbf{y}_{i,t}) - \eta d_i(t))$
\ENDFOR
\STATE Return $\mathbf{w}_i(T)$.
\end{algorithmic}
\end{algorithm}
\section{Analytical Results}
In this section, we rigorously analyze the properties of \sys{}. We fully introduce all assumptions, analyze the convergence behavior of \sys{}, and finally compare \sys{} to previous work. To improve readability, Table \ref{tab:notation} summarizes the meaning of each piece of notation.
\begin{table}[h!]
    \caption{Notation}
    \centering
    \resizebox{\linewidth}{\height} {
    \begin{tabular}{|c|l|}
        \hline
        Symbol & Meaning\\
        \hline
         $t$ & Iteration  \\
         $T$ & Total number of iterations\\ 
         $E_t$ & Connectivity graph at iteration $t$\\
         $P(t)$ & Averaging matrix\\
         $\zeta, \kappa, \eta$ & Constants related to the graph connectivity\\         
         $\phi(\cdot)$ & Function satisfying $\nabla \phi(x) = h(x)$\\
         $h$ & Used in protocol as mapping function\\
         $D_{\phi}(\cdot, \cdot)$ & Bregman divergence associated with $\phi$\\
         $r$ & Degree of uniform convexity of $\phi$\\
         $\sigma$ & Coefficient of convexity of $\phi$\\
         $f_i(\cdot)$ & Local loss function at node $i$\\
         $F(\cdot)$ & Global loss function\\
         $G_l$ & Upper bound on gradient of $f_i$\\
         $\textbf{w}_{i,t}$ & Value held at node $i$ at iteration $t$\\
         $\textbf{y}_{i,t}$ & Intermediate average\\
         $m$ & Number of devices\\
         $\eta$ & Learning rate\\
         \hline
    \end{tabular}
    }
    \label{tab:notation}
\end{table}

\label{sec:convergences}
\subsection{Assumptions}
\begin{assumption}[$h$ as the gradient of another function]
\label{ass:isMD}
There exists a function $\phi$ such that $\nabla \phi(x) = h(x)$ for all $x$. 
\end{assumption}
In section 3, we use $h(x)$ to specify the framework. As we'll see,
it is more natural to express the analytical results using $\phi$ instead of $h$.

\begin{assumption}[Connectivity]
\label{ass:connectivity}
The network $G= (V,E_t)$ and the connectivity weight matrix $P(t)$ satisfy the following:
\begin{itemize}
    \item $P(t)$ is doubly stochastic for all $t\geq 1$; that is $\sum_{j=1}^m [P(t)]_{ij}=1$ and $\sum_{i=1}^m [P(t)]_{ij}=1$, for all $i,j\in V$.
    \item There exists a scale $\zeta>0$, such that $[P(t)]_{ii}\geq \zeta$ for all $i$ and $t\geq 1$, and $[P(t)]_{ij}\geq \zeta$, if $\{i,j\}\in E_t$.
    \item There exists an integer $B\geq 1$ such that the graph $(V, E_{kB+1}\cup \cdots \cup E_{(k+1)B})$ is strongly connected for all $k\geq 0$.
\end{itemize}
\end{assumption}

\begin{definition}[Bregman Divergence]
The Bregman Divergence of a function $\phi$ is defined as
\begin{equation}
    D_{\phi}(x,y) = w(x)-w(y)-\langle \nabla w(y), x-y \rangle.
\end{equation}
Note that the Bregman divergence is defined with respect to the function $\phi$.
\end{definition}

\begin{definition} [Uniform Convexity]
Consider a differentiable convex function $\varphi: \RR^d \to \RR$, an exponent $r\geq 2$, and a constant $\sigma >0$. Then, $\varphi$ is $(\sigma, r)$-uniformly convex with respect to a $\lVert \cdot \rVert$ norm if for any $x,y \in \RR^d $,
\begin{equation}
  \varphi(x)\geq \varphi(y) + \langle \nabla \varphi(y), x-y\rangle + \frac{\sigma}{r} \lVert x-y\rVert^r.  
\end{equation}
Note that for $r=2$, this is know as \textit{strong convexity}. This assumption also implies that $D_{\varphi}(x,y) \geq \frac{\sigma}{r}\lVert x-y\rVert^r$.
\end{definition}

\begin{assumption}[Bounded Gradient]
\label{ass:boundedgradient}
Functions $f_{i}$ for $i\in [m]$ are convex. Additionally, all $f_i$ are $G_l$-Lipschitz \cite{yuan2020distributed}.
This implies that for all $x,y$ pairs
\begin{equation}
    \lVert f_i(x) - f_i(y)\rVert \leq G_l \lVert x-y\rVert.
\end{equation}
As a corollary, for all $x$, it is true that $\lVert \nabla f_{i} (x) \rVert \leq G_l$.
\end{assumption}
While relatively strong, this assumption is commonly used in DML, and is used in \cite{verbraeken2020survey, bubeckReview,srebro2011, yuan2020distributed, nedic2008distributed}. In our analysis, it is used to bound the distance between local models.

\subsection{Proof under Uniform Convexity}
To aid analysis, introduce the sequence $\textbf{y}_{i,t}$ satisfying $\nabla \phi(\textbf{y}_{i,t})= \sum_{j=1}^m P(t)_{ij}  \nabla \phi(\textbf{w}_{j,t}) $. We first summarize the properties of Algorithm \ref{alg:new} with the following lemma.
\begin{lemma}
The $\textbf{w}_{i,t}$ and $\textbf{y}_{i,t}$ sequences satisfy
\begin{align}
\nabla \phi(\textbf{y}_{i,t}) &= \sum_{j=1}^m P(t)_{ij} \nabla \phi(\textbf{w}_{j,t}),\\
\nabla \phi(\textbf{w}_{i,t+1}) 
&= \nabla \phi(\textbf{y}_{i,t}) - \eta \nabla f_i(\textbf{w}_{i,t}).\label{eq:gradphi}
\end{align}
Also, define
\[\overline{\textbf{w}_t} = [\nabla \phi]^{-1}(\frac1m \sum_{i=1}^m \nabla \phi(\textbf{w}_{i,t})) =[\nabla \phi]^{-1}(\frac1m \sum_{i=1}^m \nabla \phi(\textbf{y}_{i,t})).\]
\end{lemma}
\begin{proof}
The first part summarizes the update rules in Algorithm \ref{alg:new} implicitly. The second part is self-consistent because $P(t)$ is doubly stochastic.
\end{proof}
Using the Connectivity Assumption (Assumption \ref{ass:connectivity}),  we can bound the distance between local models held at different devices.
\begin{lemma}
\label{lemma:consensus}
Under Assumption \ref{ass:connectivity}, for each device $i$, after $t$ iterations:
\begin{equation}
    \lVert \nabla \phi(\textbf{w}_{i,t}) - \nabla \phi(\overline{\textbf{w}_t}) \rVert \leq \vartheta( \kappa^{t-1} \sum_{j=1}^m \lVert \nabla \phi(\textbf{w}_{j,0})\rVert +  \frac{m\eta G_l}{1-\kappa} + 2\eta G_l),
\end{equation}
where $\vartheta = \left(1-\frac{\zeta}{4m^2} \right)^{-2}$ and $\kappa = \left(1-\frac{\zeta}{4m^2} \right)^{\frac1B}$.
\end{lemma}
\begin{proof}
Define the matrix $P(t,s) = P(t)P(t-1)\cdots P(s+1)P(s)$. Then, under Assumption \ref{ass:connectivity}, Corollary 1 in \cite{nedic2008distributed} states that
\begin{equation}
\label{eq:expmix}
|P(t,\tau)_{ij} - \frac1m | \leq \vartheta \kappa^{t-\tau}.
\end{equation}
where $\vartheta$ and $\kappa$ are defined in the lemma statement. The rest of the proof can be found in the Appendix.
\end{proof}

\begin{theorem}[Convergence Behavior]
\label{thm:maintheorem}
Consider a $(\sigma,r)$ uniformly convex $\phi$ and the sequence $\textbf{w}_{i,t}$ under Algorithm \ref{alg:new} with constant step size $\eta$. Then, under Assumptions \ref{ass:isMD}, \ref{ass:connectivity}, and \ref{ass:boundedgradient}, if $x^{\ast}$ is the value that minimizes $F(\textbf{w}) = \sum_{i=1}^m f_i(\textbf{w})$, then
\begin{equation}
\begin{aligned}
    & \min_t F(\overline{\textbf{w}_t}) - F(x^{\ast}) \\
    & \leq \frac{2m G_l}{T}  \sum_{t=0}^{T-1} \sqrt[r-1]{\frac{\vartheta}{\sigma}  ( \kappa^{t-1} \sum_{j=1}^m \lVert \nabla \phi(\textbf{w}_{j,0})\rVert +  \frac{m\eta G_l}{1-\kappa}+2\eta G_l) }\\ &+m\cdot \frac{r-1}{r}  \sqrt[r-1]{\frac{1}{\sigma }\cdot \eta \cdot G_l^r}+ \frac{m\cdot D_{\phi}(x^{\ast},\overline{\textbf{w}_0})}{\eta T}.
\end{aligned}    
\end{equation}

\end{theorem}
\begin{proof}
See Appendix Section \ref{subsec:theorem1}. 
\end{proof}
In this theorem (Theorem \ref{thm:maintheorem}), the first term uses Lemma \ref{lemma:consensus}, and is caused by the differences between local models held at different devices. This is the effect of the \textit{diverging forces} mentioned in earlier sections. The second term represents the error caused by a non-zero learning rate in the Mirror Descent process, and the third term represents the lingering effects of the initialization.

\begin{corollary}
The bound on \sys{}'s (Algorithm \ref{alg:new}) loss is $O(m \sqrt[r-1]{\eta m} +\frac{m}{\eta T})$.
\end{corollary}
\begin{proof}
The first term becomes $O(m\cdot \sqrt[r-1]{m \eta})$ because since $\kappa <1$, the $\kappa^t$ term goes to 0 as $t$ gets large. The second term is $O(m\sqrt[r-1]{\eta})$, but this is smaller than the first term so we drop it. The final term is clearly $O(\frac{m}{\eta T})$.
\end{proof}
\begin{corollary}
For $\eta=O(\sqrt[r] {\frac{T^{r-1}}{m}})$, the upper bound on the loss becomes $O(m \sqrt[r-1]{\eta m} +\frac{m}{\eta T}) = O(\sqrt[r]{\frac{m^{r+1}}{T}})$.
\end{corollary}

\subsection{Comparison to Previous Work}
Table \ref{tab:convrate} summarizes the big $O()$ notation convergence rates of $f(\overline{\textbf{w}_t})- f^{\ast}$. Each previous work also assumes bounded gradients. Our analysis recovers the same bounds as the state-of-the-art in both DML and single device mirror descent with generic uniform convexity assumptions.
\begin{table}[h!]
    \centering
   \caption{Convergence Rates in terms of $\eta$, $r$, $m$, and $T$}
    \resizebox{\linewidth}{12mm}{
    \begin{tabular}{|c|c|c|c|}
    \hline
    Paper & $f$ error & with optimal $\eta$ & Recovered\\
    \hline
    \sys{} & $O(m \sqrt[r-1]{\eta m} +\frac{m}{\eta T})$ & $O(\sqrt[r]{\frac{m^{r+1}}{T}})$ & .\\ 
     Bubeck \cite{bubeckReview} $(m=1,r=2)$ & $O(\eta + \frac1{\eta T} )$ & $O(\frac{1}{\sqrt{T}})$ & Yes\\
     Srebro~\cite{srebro2011} $(m=1)$ & $O(\eta^{\frac1{r-1}} + \frac1{\eta T} )$ & $O(\frac{1}{\sqrt[r]{T}})$ & Yes\\
    Nedic~\cite{nedic2009distributed} $(GD \to r=2)$ & $O(\eta m^2+ \frac{m}{\eta T})$ & $O(\frac{m\sqrt{m}}{\sqrt{T}})$ & Yes \\
     Yuan~\cite{yuan2020distributed} ($r=2$ MD) & $O(\eta m^2+ \frac{m}{\eta T})$& $O(\frac{m\sqrt{m}}{\sqrt{T}})$ & Yes\\     
     \hline    
    \end{tabular}
     }
    \label{tab:convrate}
\end{table}

Nedic~\cite{nedic2009distributed} is standard gradient descent, and thus has $\phi(x)=\frac12 \lVert x\rVert^2$, which is $(1,2)$-uniformly convex. Yuan\cite{yuan2018} considers distributed mirror descent under strong convexity $(r=2)$, and their stated bound is $O(\eta m+ \frac{m}{\eta T})$. However, based on their analysis, the bound should be $O(\eta m^2+ \frac{m}{\eta T})$.

\subsection{Verification of $h(x)=x^p$}
In order to turn the \sys{} protocol into a usable method, one needs to select a specific $h$ function. In our experiments, we use $\phi(x) = \frac1{p+1} \lVert x\rVert^{p+1}$ which gives $h(x) = \nabla \phi(x) = x^p$. Such functions satisfy the Uniform Convexity condition due to Proposition \ref{prop:power}. 
\begin{proposition}[Uniform Convexity of Power Functions]
\label{prop:power}
For $p\geq 1$, the function $\varphi_p(x) = \frac1{p+1} \lVert x\rVert^{p+1}$ is uniformly convex with degree $p+1$. This is because $\lVert \nabla \varphi_p(x)-\nabla \varphi_p(y)\rVert = \lVert x^p - y^p \rVert\geq \frac{1}{2^{p-1}}\lVert x-y \rVert$, and as a corollary we have
\begin{equation}
    D_{\varphi_p}(x,y) \geq \frac{1}{2^{p-1}} \cdot \frac{1}{p+1}\lVert x-y\rVert^{p+1}.
\end{equation}
\end{proposition}
\begin{proof}
A proof can be found on page 7 of \cite{Doikov_2021}.
\end{proof}
This proposition shows that $h(x)= x^p$ satisfies the necessary assumptions in Theorem \ref{thm:maintheorem}, and with this $h$ function \sys{} instructs local models to take a \textit{weighted power mean} of received models. This is a generalization of traditional linear aggregation and using $h(x)=x^p$ functions, one can hit any integer value of $r$ with $h(x) = x^{r-1}$ since $r=p+1$.

\section{Experimental Evaluation}
\label{sec:evaluationss}
In this section, we empirically evaluate the performance of \sys{} under Non-IID datasets and dynamic/sparse network topologies. We also examine scalability with respect to $m$, the number of devices.

\subsection{Experimental Methodology}
\label{subsec:settings}
The experimental platform is composed of 8 Nvidia Tesla T4 GPUs, 16 Intel XEON CPUs and 256GB memory. We construct an experimental environment based on 
Ray~\cite{ray2018}, which is an open-source framework that provides a universal API for building high performance distributed applications.

\para{Weak-Connectivity Setup.}
\label{subsubsec:topology}
We use the Extendable Mobile Ad-hoc Network Emulator (EMANE)~\cite{chu2022model} platform to simulate a real low-connectivity environment with 5-20 distributed devices. Each device is configured with WiFi 802.1a and communicates with other devices in the Ad-Hoc mode. We use $density = \frac{n}{N}$ to measure the network density of the topology, where $n$ is the number of available tunnels, and $N$ is the total number of tunnels.

\para{Models and Datasets.}
We use a convex model Logistic Regression (LR) \cite{hosmer2013applied}. 
We conduct our experiments using a public dataset MNIST \cite{lecun1998gradient}. 

\para{Non-IID Dataset Partition Setup.}
\label{subsubsec:noniid_data_partition}
To implement Non-IID data, we adopt $\alpha$ as a degree controller of disjoint Non-IID data under a Dirichlet distribution~\cite{lin2021quasi}, and the dataset for each device is pre-assigned before training. 
For small $\alpha$, devices will hold samples from fewer classes. $\alpha=0.1$ is the extreme scenario in which each local device holds samples from only one class, while $\alpha=10$ is the IID scenario.

\begin{table*}[htbp]
    \scriptsize
    \centering
    \caption{Performance of Different p Value}
    \resizebox{2\columnwidth}{!}{
    \begin{tabular}{|m{1cm}|m{0.9cm}|m{0.9cm}|m{0.9cm}|m{0.9cm}|m{0.9cm}|m{0.9cm}|m{0.9cm}|m{0.9cm}|m{0.9cm}|m{0.9cm}|m{0.9cm}|m{0.9cm}|m{0.9cm}|m{0.9cm}|m{0.9cm}|} 
    \hline
    \multirow{2}{*}{p Value} & \multicolumn{15}{c|}{0 - 200 Iterations} \\
    \cline{2-16}
    & \makecell[c]{p=1} & \makecell[c]{p=2} & \makecell[c]{p=3} & \makecell[c]{p=4} & \makecell[c]{p=5} & \makecell[c]{p=6} & \makecell[c]{p=7} & \makecell[c]{p=8} & \makecell[c]{p=9} & \makecell[c]{p=10} & \makecell[c]{p=11} & \makecell[c]{p=12} & \makecell[c]{p=13} & \makecell[c]{p=14} & \textbf{\makecell[c]{p=15}} \\
    \hline
    \makecell[c]{Avg Acc} & 
    \makecell[c]{68.48} &
    \makecell[c]{68.09\\(-0.57\%)} &
    \makecell[c]{70.92\\(3.56\%)} &
    \makecell[c]{71.47\\(4.37\%)} &
    \makecell[c]{76.38\\(11.54\%)} &
    \makecell[c]{72.59\\(6.0\%)} &
    \makecell[c]{79.07\\(15.46\%)} &
    \makecell[c]{73.28\\(7.01\%)} &
    \makecell[c]{79.94\\(16.73\%)} &
    \makecell[c]{73.96\\(8.0\%)} &
    \makecell[c]{80.43\\(17.45\%)} &
    \makecell[c]{74.45\\(8.72\%)} &
    \makecell[c]{80.75\\(17.92\%)} &
    \makecell[c]{74.78\\(9.2\%)} &
    \textbf{\makecell[c]{81.28\\(18.69\%)}} \\
    \hline
    \makecell[c]{Min Loss} & 
    \makecell[c]{1.6763} &
    \makecell[c]{1.7394\\(3.76\%)} &
    \makecell[c]{1.666\\(-0.61\%)} &
    \makecell[c]{1.6508\\(-1.52\%)} &
    \makecell[c]{1.5393\\(-8.17\%)} &
    \makecell[c]{1.6169\\(-3.54\%)} &
    \makecell[c]{1.4946\\(-10.84\%)} &
    \makecell[c]{1.5905\\(-5.12\%)} &
    \makecell[c]{1.481\\(-11.65\%)} &
    \makecell[c]{1.5679\\(-6.47\%)} &
    \makecell[c]{1.4684\\(-12.4\%)} &
    \makecell[c]{1.557\\(-7.12\%)} &
    \makecell[c]{1.4639\\(-12.67\%)} &
    \makecell[c]{1.5435\\(-7.92\%)} &
    \textbf{\makecell[c]{1.4563\\(-13.12\%)}} \\
    \hline
    \makecell[c]{Converge\\Iterations} & 
    \makecell[c]{115} &
    \makecell[c]{161\\(40.0\%)} &
    \makecell[c]{187\\(62.61\%)} &
    \makecell[c]{117\\(1.74\%)} &
    \makecell[c]{94\\(-18.26\%)} &
    \makecell[c]{101\\(-12.17\%)} &
    \makecell[c]{77\\(-33.04\%)} &
    \makecell[c]{94\\(-18.26\%)} &
    \makecell[c]{70\\(-39.13\%)} &
    \makecell[c]{94\\(-18.26\%)} &
    \makecell[c]{68\\(-40.87\%)} &
    \makecell[c]{94\\(-18.26\%)} &
    \makecell[c]{68\\(-40.87\%)} &
    \makecell[c]{81\\(-29.57\%)} &
    \textbf{\makecell[c]{50\\(-56.52\%)}} \\
    \hline
    \end{tabular}}
    \label{tab:different p compare table}
\end{table*}

\para{{Metrics.}}
\label{subsubsec:performance_metrices}
We use three metrics to measure the performance of \sys{}
\begin{itemize}
	\item \textbf{Model Accuracy.} The accuracy measures the proportion of data correctly identified by the model. We compute the average accuracy as $\frac{1}{m} \sum_{i=1}^m \frac{\chi_i(r)}{\chi_i}$ where
	$\chi_i$ denotes the number of samples in device $i$, and $\chi_i(r)$ denotes the number of correctly identified samples at device $i$.
	\item \textbf{Test Loss.} We use the average of the test losses, where a test loss measures the difference between model outputs and observation results. 
	\item \textbf{Convergence Speed.} 
	We record the loss and iteration number at each iteration; the convergence speed depends on the number of iterations needed to achieve specific loss values.
	\end{itemize}
\begin{figure}[b]
	\centering
	\begin{minipage}[c]{0.2\textwidth}
		\centering
		\includegraphics[width=1.6in]{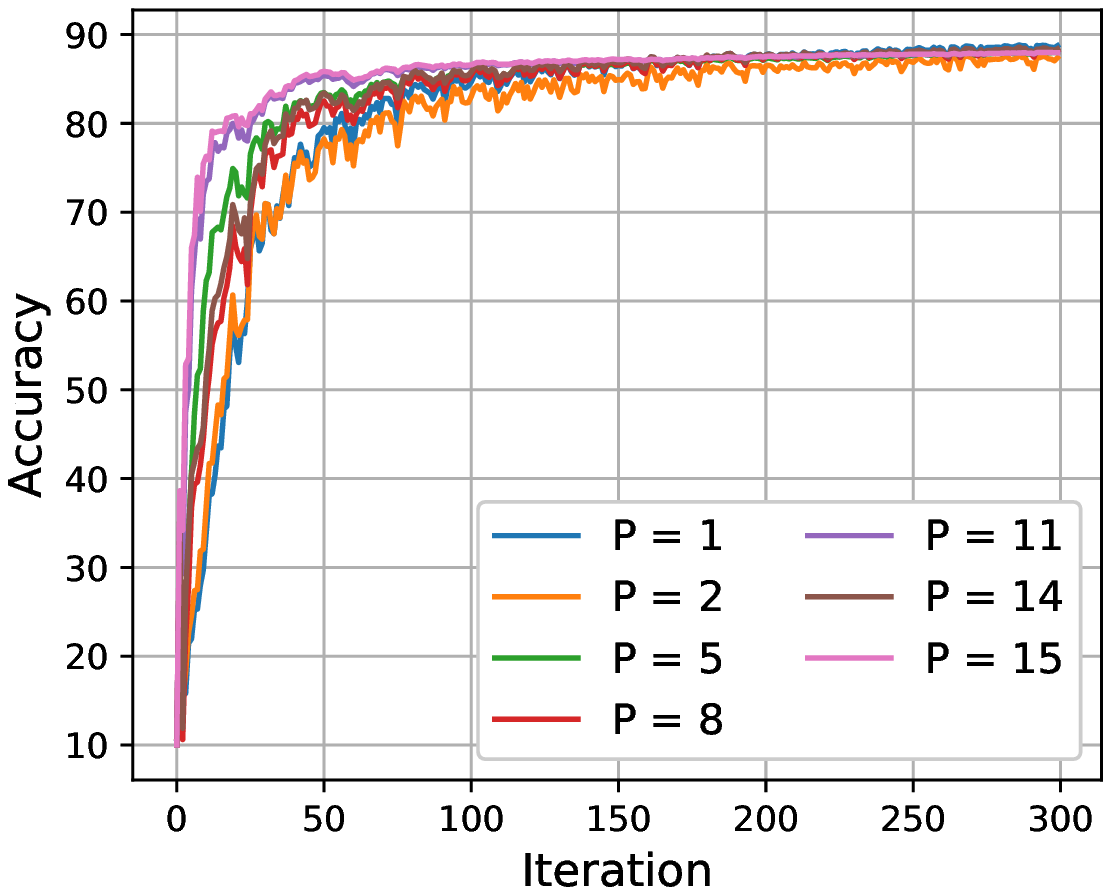}
		\centerline{(a) Accuracy}
		
	\end{minipage}
	\hspace{0.02\textwidth}
	\begin{minipage}[c]{0.2\textwidth}
		\centering
		\includegraphics[width=1.6in]{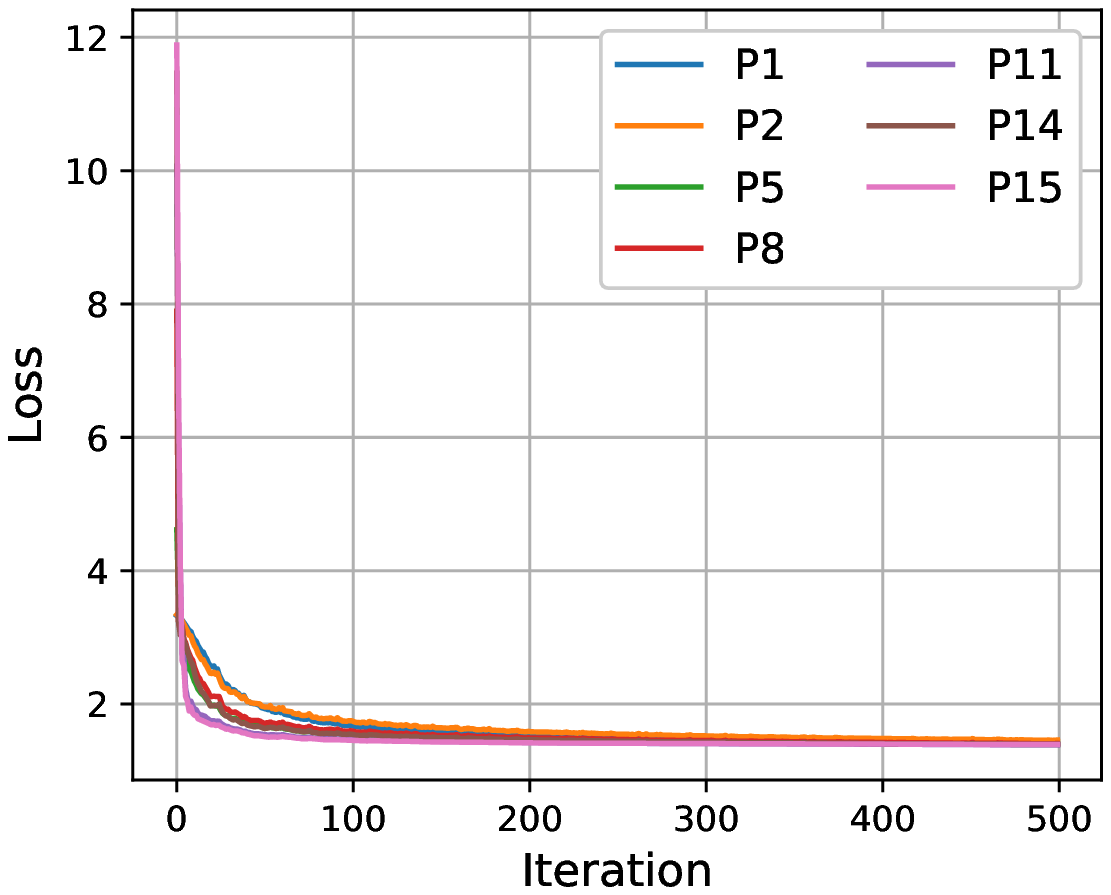}
		\centerline{(b) Loss}
	\end{minipage}
	\caption{Convergence with Different $p$ Value.}
	\label{figure:different_p}
\end{figure}

\para{Baselines.}
\label{subsubsec:Baselines_Setup}
We compare \sys{} with the traditional linear aggregation framework $(p=1)$ and SwarmSGD \cite{2019SwarmSGD} in weak-connectivity environments. In order to provide a fair comparison among all methods, we make minor adjustments for the two baselines as follows.
\begin{itemize}
    \item \textit{p=1}. We define it as a class of methods using a linear aggregation function~\cite{2017DPSGD,nedic2008distributed, neglia2020decentralized}. All of these methods are a special case of \sys{}. 
    \item \textit{SwarmSGD}. We set the number of local SGD updates equal to 1; the selected pair of devices perform only one single local SGD update before aggregation.
    \end{itemize}

\begin{figure}[t]
    \centering
    \begin{minipage}[c]{0.15\textwidth}
    	\centering
    	\includegraphics[width=1.2in]{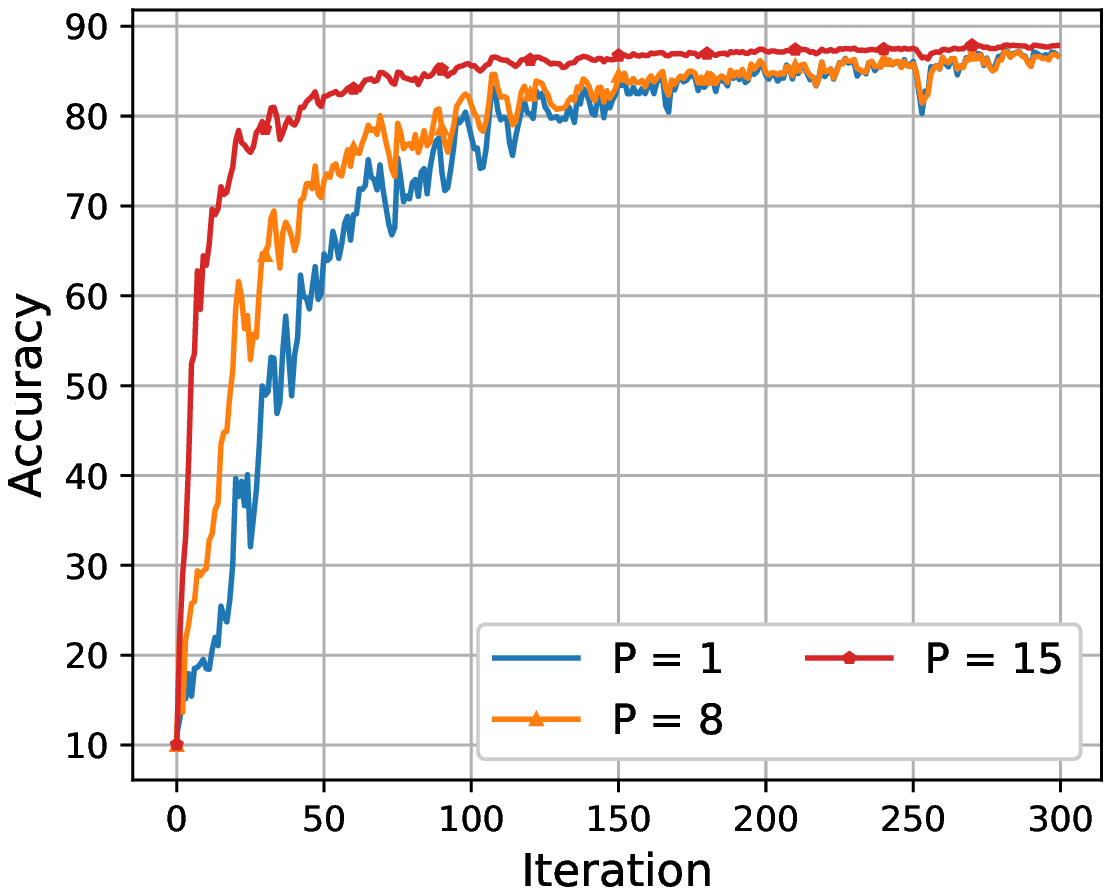}
    	\centerline{(a) Density=0.2}
    \end{minipage}
    \begin{minipage}[c]{0.15\textwidth}
    	\centering
    	\includegraphics[width=1.2in]{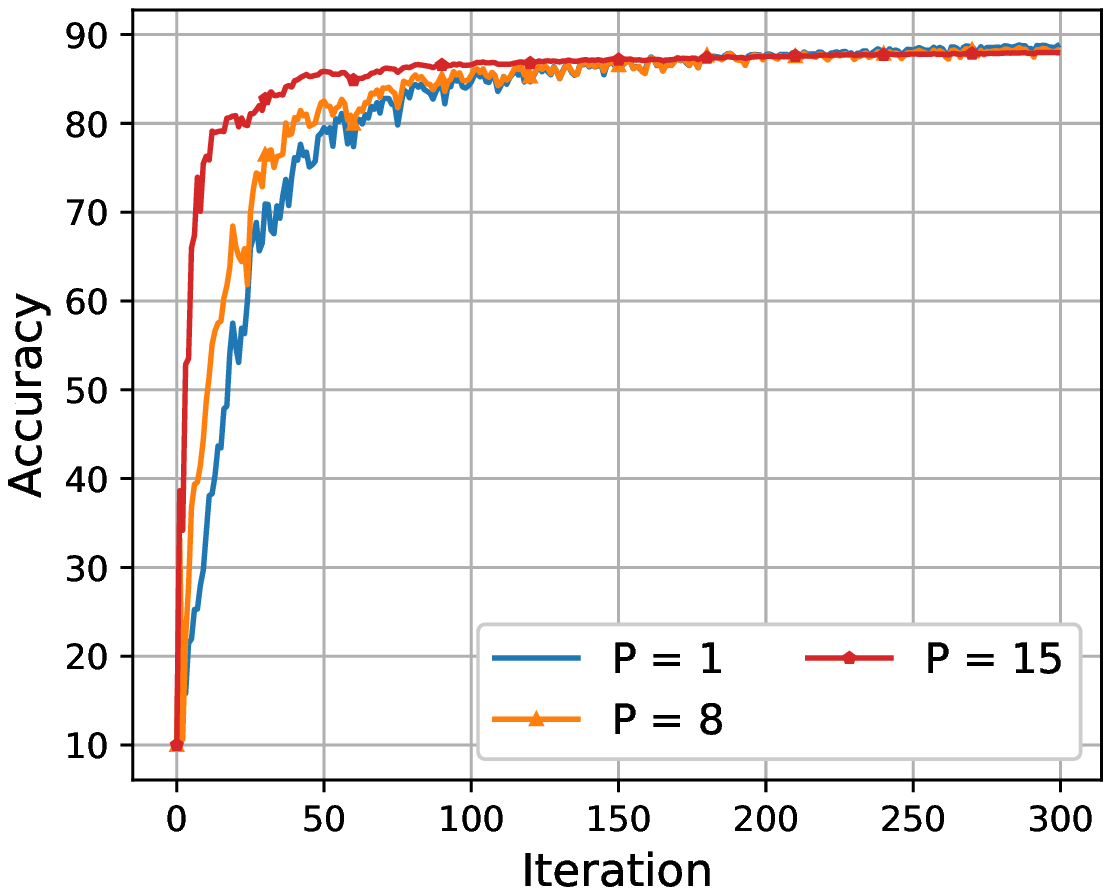}
    	\centerline{(b) Density=0.5}
    \end{minipage}
    \begin{minipage}[c]{0.15\textwidth}
    	\centering
    	\includegraphics[width=1.2in]{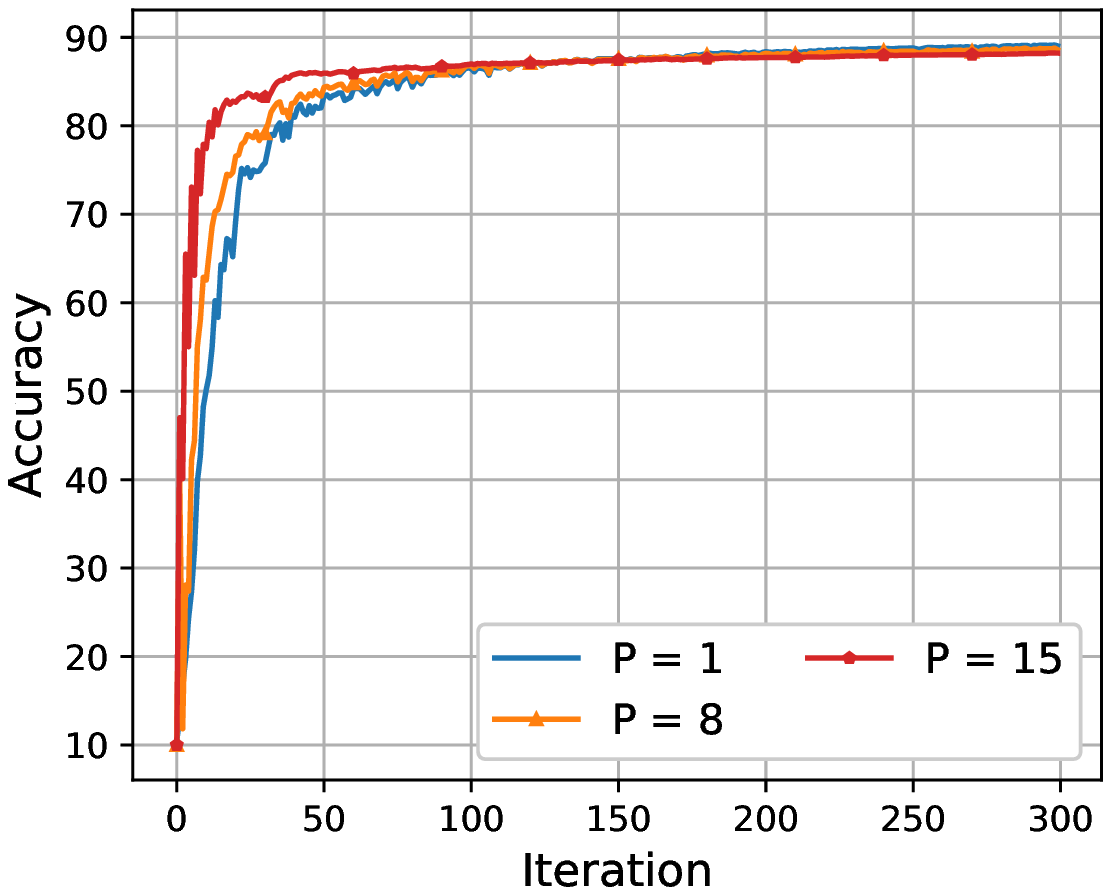}
    	\centerline{(c) Density=0.8}
    \end{minipage}
    \caption{Accuracy with Different Communication Densities}
    \label{fig:density_acc}
\end{figure}
\begin{figure}[t]
    \centering
    \begin{minipage}[c]{0.15\textwidth}
    	\centering
    	\includegraphics[width=1.2in]{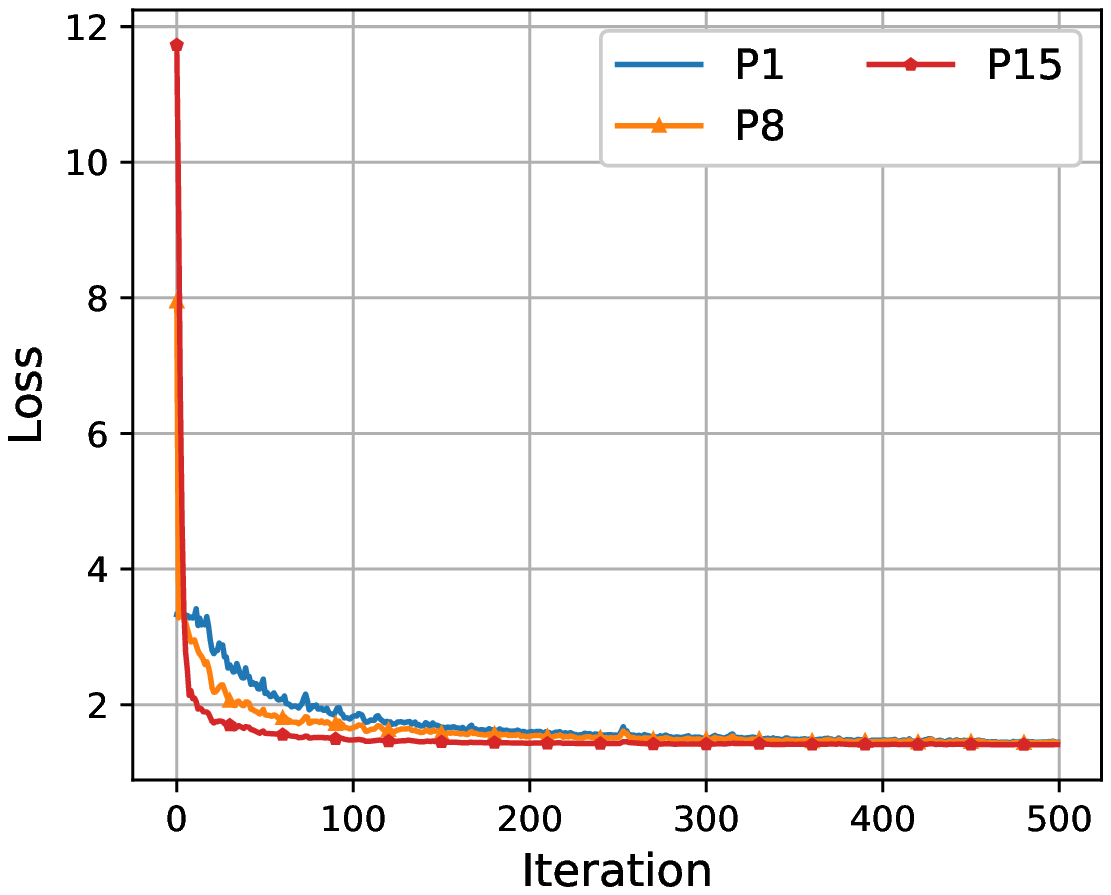}
    	\centerline{(a) Density=0.2}
    \end{minipage}
    \begin{minipage}[c]{0.15\textwidth}
    	\centering
    	\includegraphics[width=1.2in]{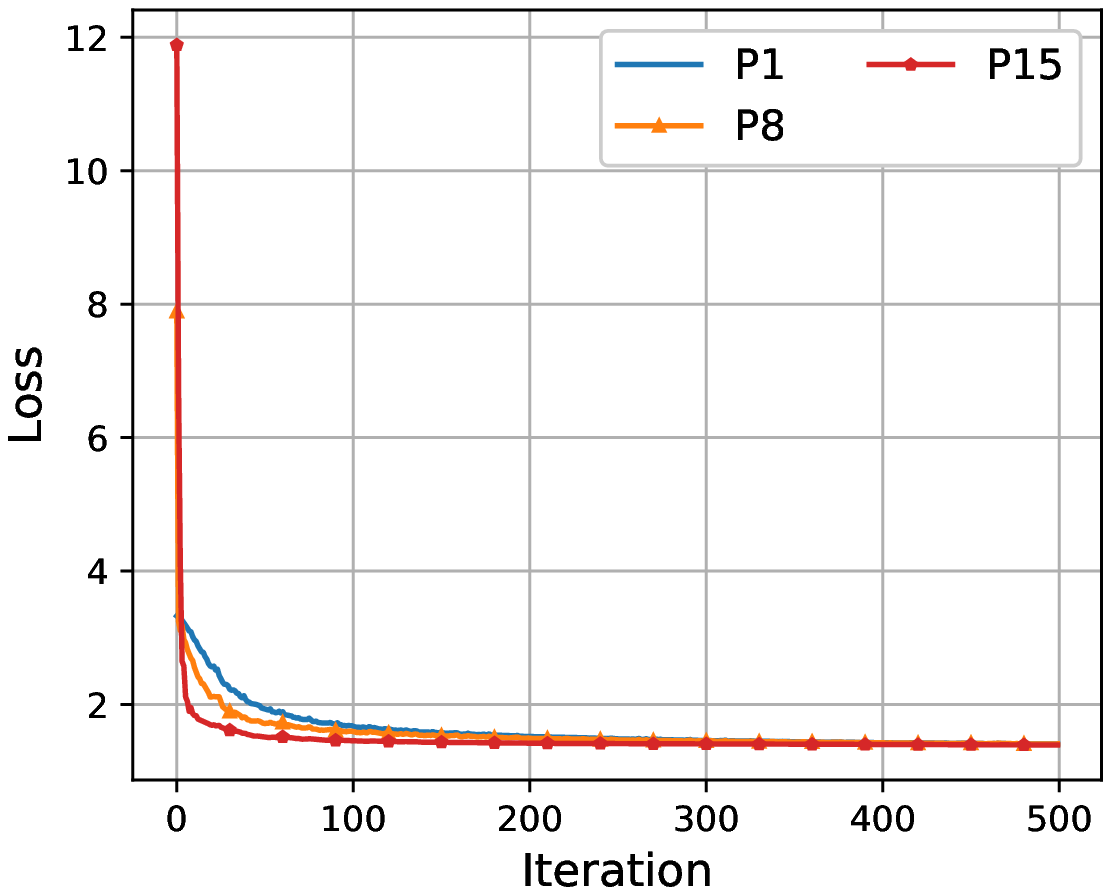}
    	\centerline{(b) Density=0.5}
    \end{minipage}
    \begin{minipage}[c]{0.15\textwidth}
    	\centering
    	\includegraphics[width=1.2in]{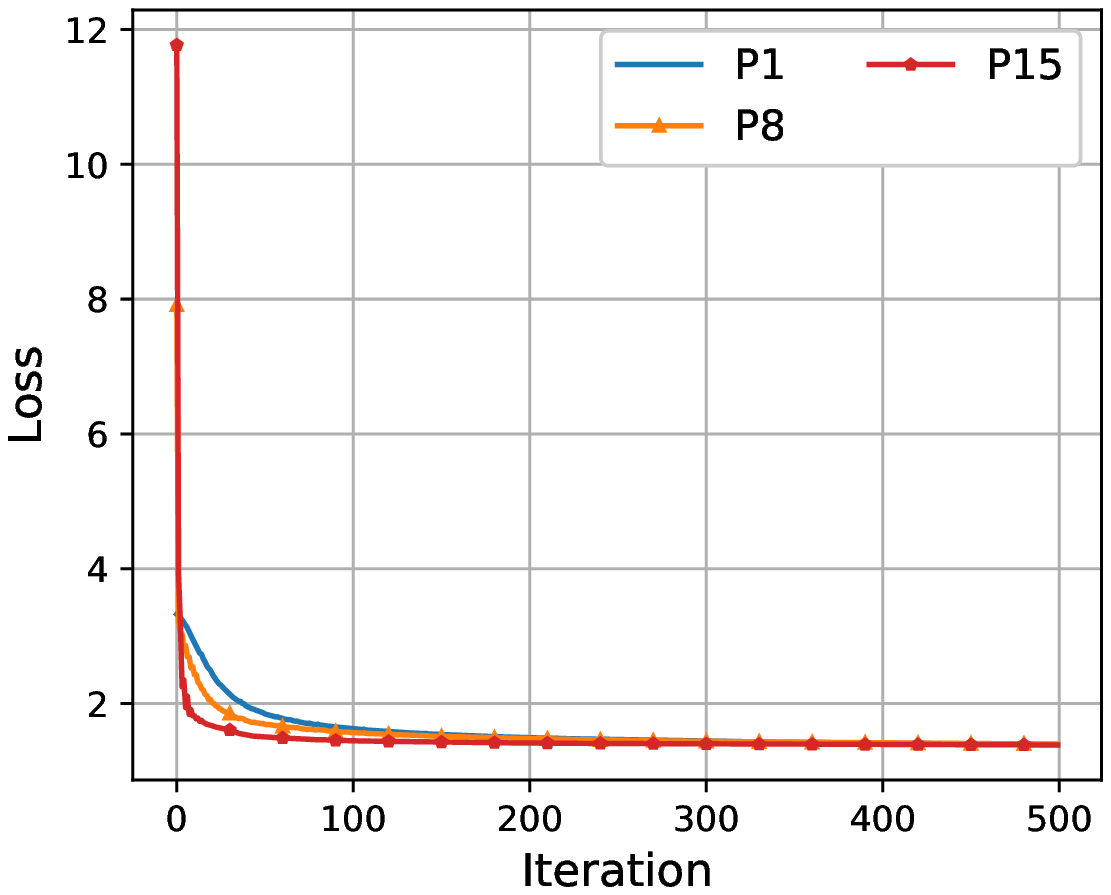}
    	\centerline{(c) Density=0.8}
    \end{minipage}
    \caption{Loss with Different Communication Densities}
    \label{fig:density_loss}
\end{figure}
\begin{figure}[t!b]
    \centering
    \begin{minipage}[c]{0.15\textwidth}
    	\centering
    	\includegraphics[width=1.2in]{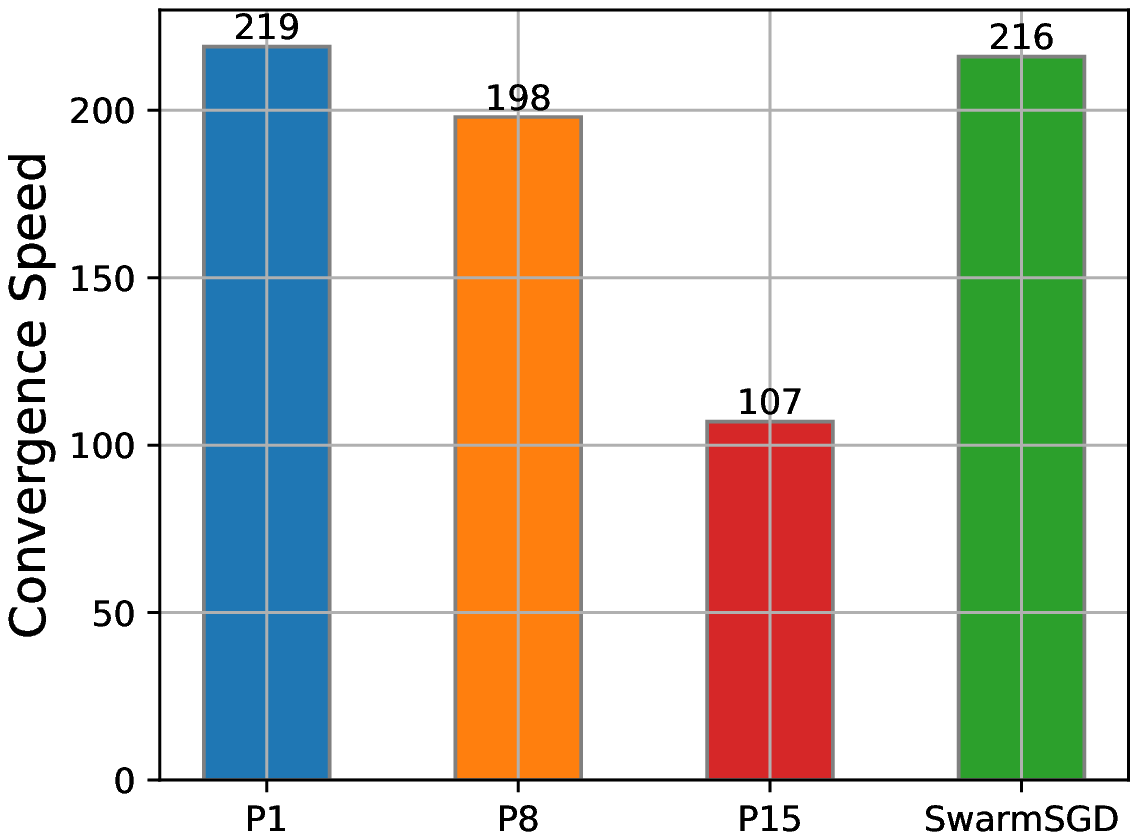}
    	\centerline{(a) Density=0.2}
    \end{minipage}
    \begin{minipage}[c]{0.15\textwidth}
    	\centering
    	\includegraphics[width=1.2in]{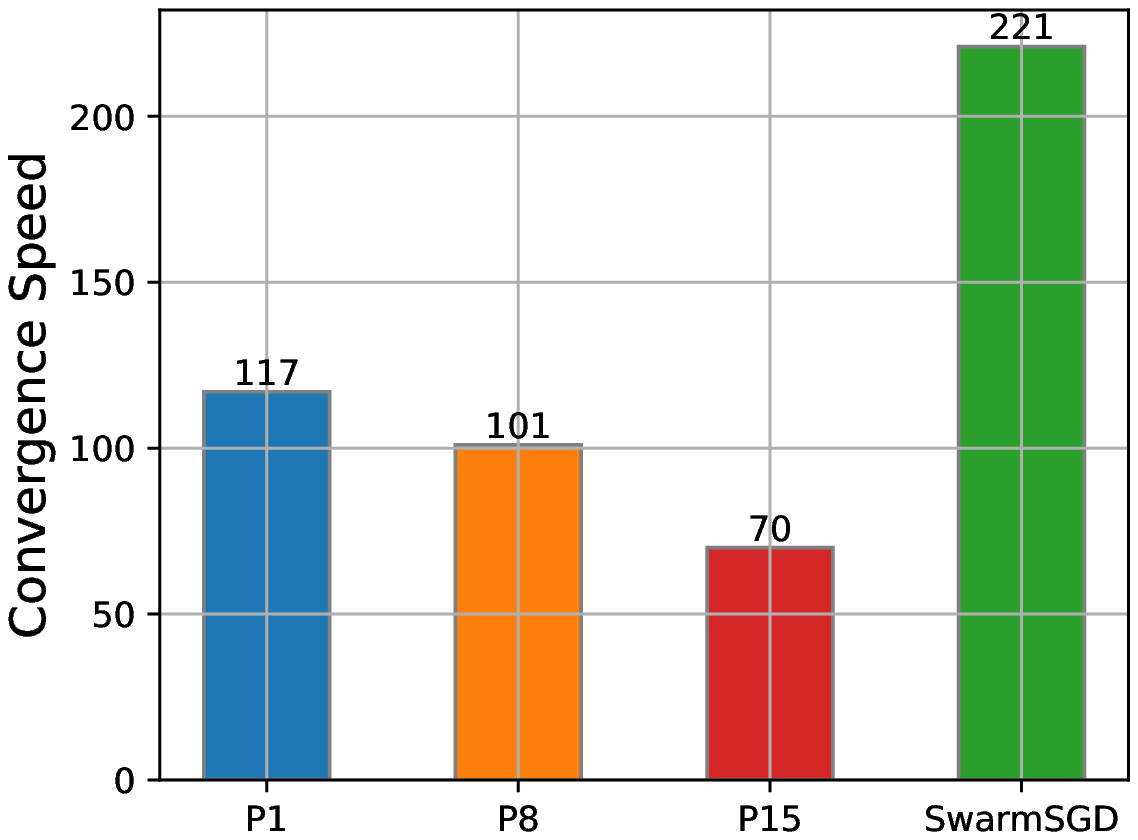}
    	\centerline{(b) Density=0.5}
    \end{minipage}
    \begin{minipage}[c]{0.15\textwidth}
    	\centering
    	\includegraphics[width=1.2in]{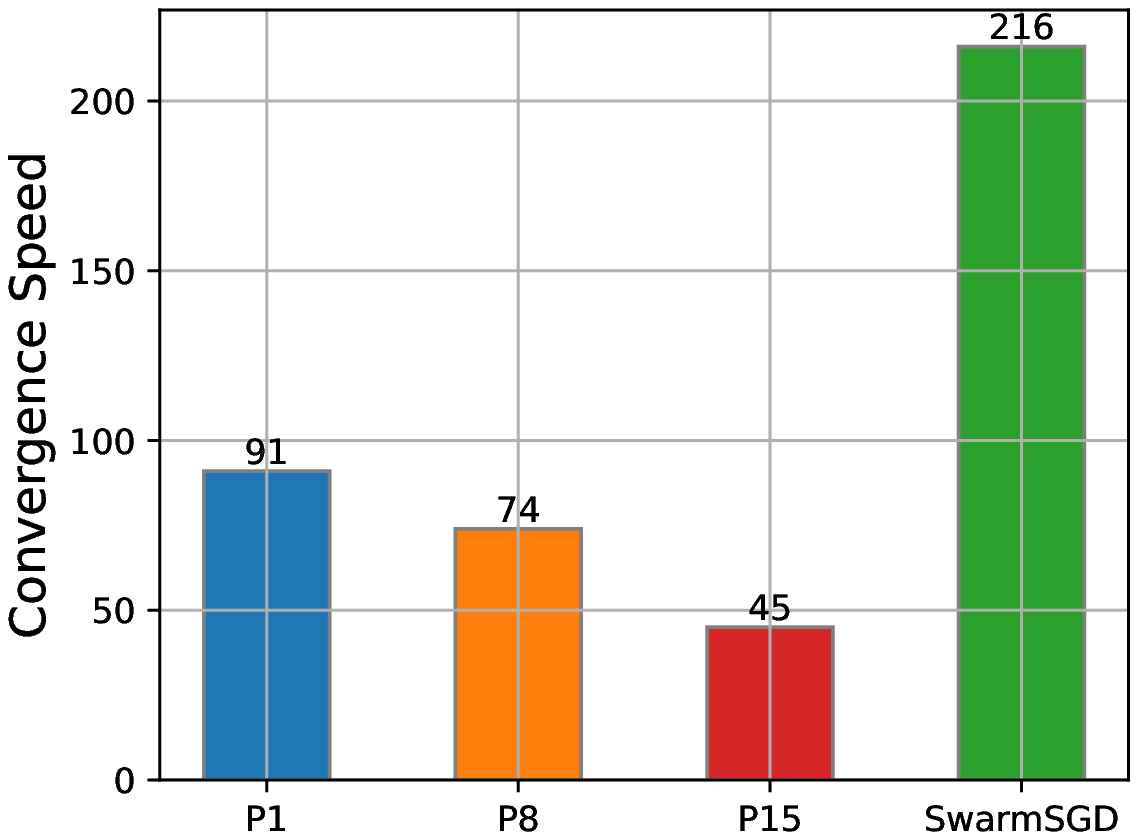}
    	\centerline{(c) Density=0.8}
    \end{minipage}
    \caption{Convergence Iterations with Different Communication Densities}
    \label{fig:density_converge}
\end{figure}

\subsection{Experiment Results}
\para{Convergence with Different p Values.}
\label{subsubsec:different_P}
First, we investigate how different $p$ values affect the performance. We train a LR model on MNIST using 10 devices with $p \in \{1, 2, ..., 15\}$. 
The results are in Figure \ref{figure:different_p} and the more detailed Table \ref{tab:different p compare table}. 

We find that the the performance of \sys{} peaks at $p=15$, and then gradually worsens as $p$ is increased further. Beyond that, large values of $p$ results in extremely small parameters after mapping and performance decreases. Therefore $p=15$ is the largest value of $p$ used in our experiments.

From Figure \ref{figure:different_p} and Table \ref{tab:different p compare table}, we can see that as $p$ increases, \sys{} achieved better performance in both model metrics and convergence speed. With the extreme Non-IID and low communication setting, \sys{} with $p=15$, on average, increased model performance by 16\% and accelerated convergence speed by up to 57\% compared to $p=1$.

\para{Convergence with Different Communication Densities.}
\label{subsubsec:different_densities}
We analyze the performance of \sys{} under different communication densities. We set $density=\{0.2, 0.5, 0.8\}$ to generate different low-communication environments. The results are illustrated in Figure \ref{fig:density_acc},\ref{fig:density_loss},\ref{fig:density_converge}, 
respectively.

The results show that \sys{} is strongly robust in low-communication environments. \sys{}  has advantages in both model performance and convergence speed in all selected $density$ values. The benefits are largest in extremely low-communication conditions; for \sys{} with $p=15$, its loss only increases by 9.5\% when the density is changed from $density=0.8$ to $density=0.2$, whereas other baselines' loss 
increase by up to 26.36\%.

\begin{figure}[t]
    \centering
    \begin{minipage}[c]{0.15\textwidth}
    	\centering
    	\includegraphics[width=1.2in]{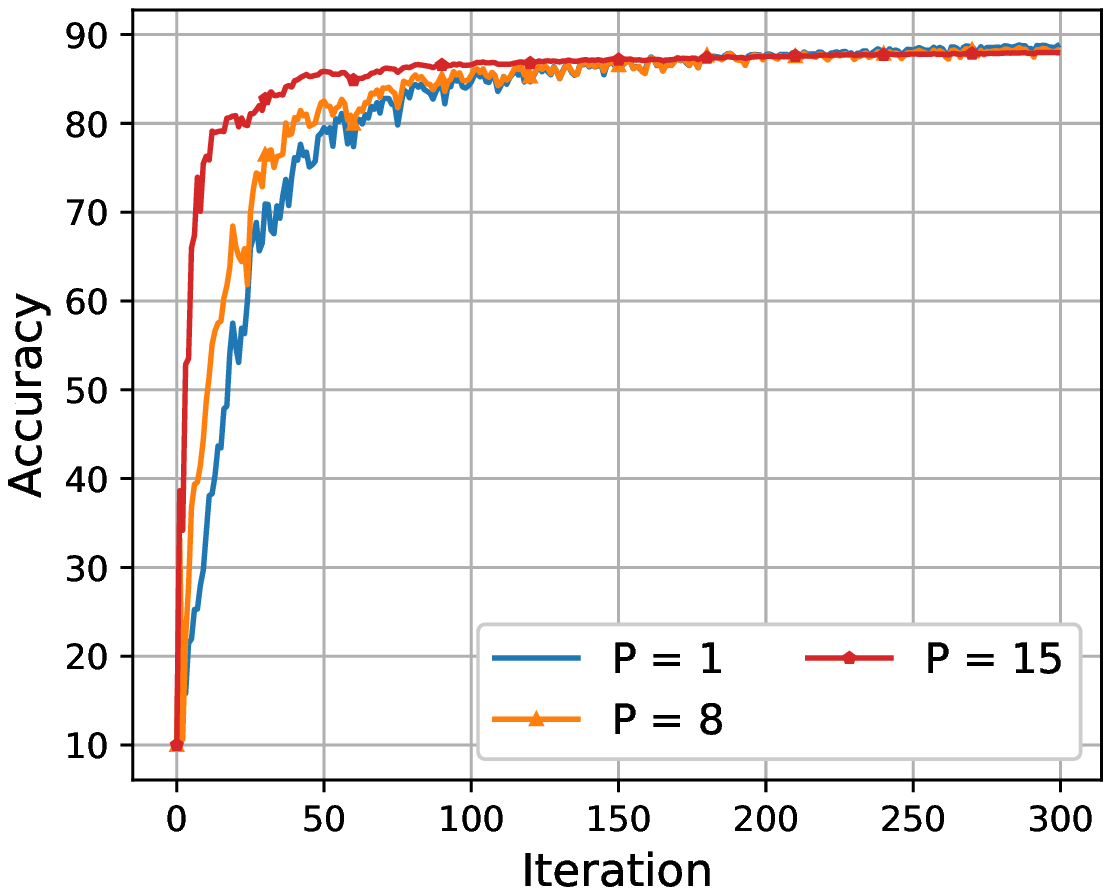}
    	\centerline{(a) Dirichlet=0.1}
    \end{minipage}
    \begin{minipage}[c]{0.15\textwidth}
    	\centering
    	\includegraphics[width=1.2in]{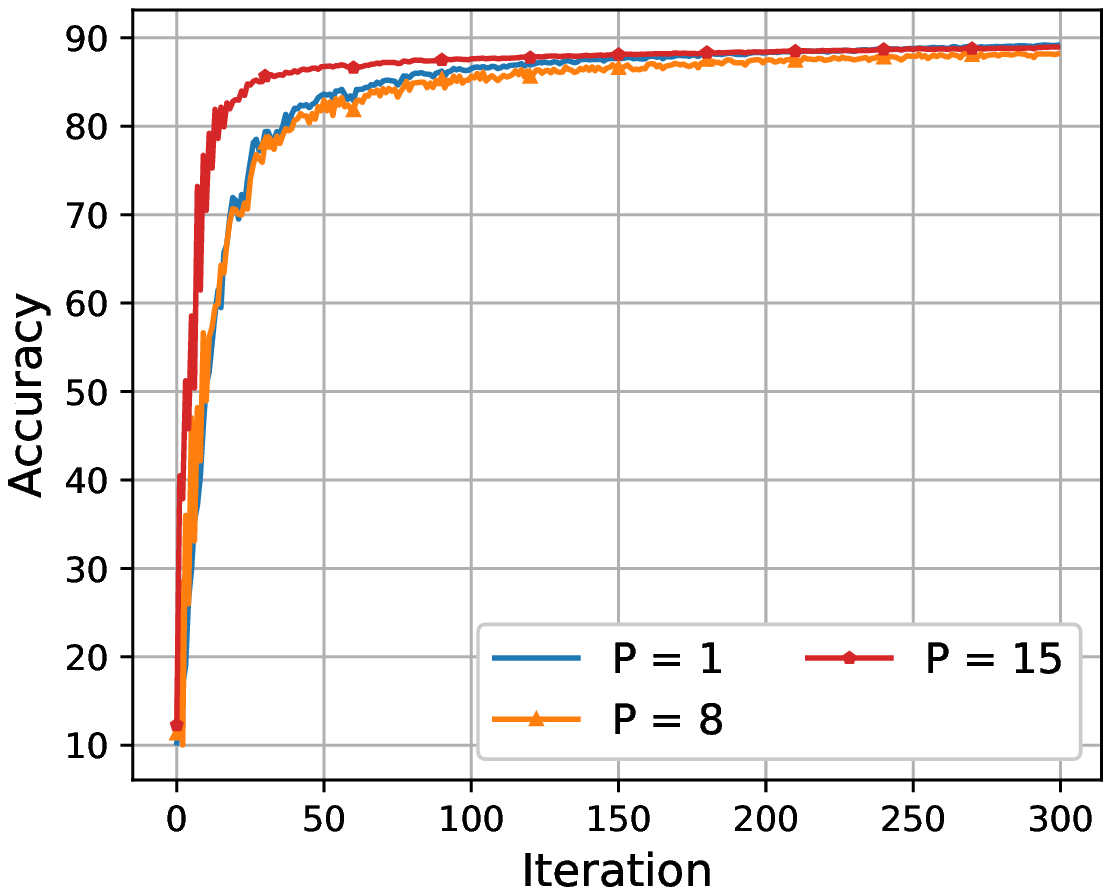}
    	\centerline{(b) Dirichlet=1}
    \end{minipage}
    \begin{minipage}[c]{0.15\textwidth}
    	\centering
    	\includegraphics[width=1.2in]{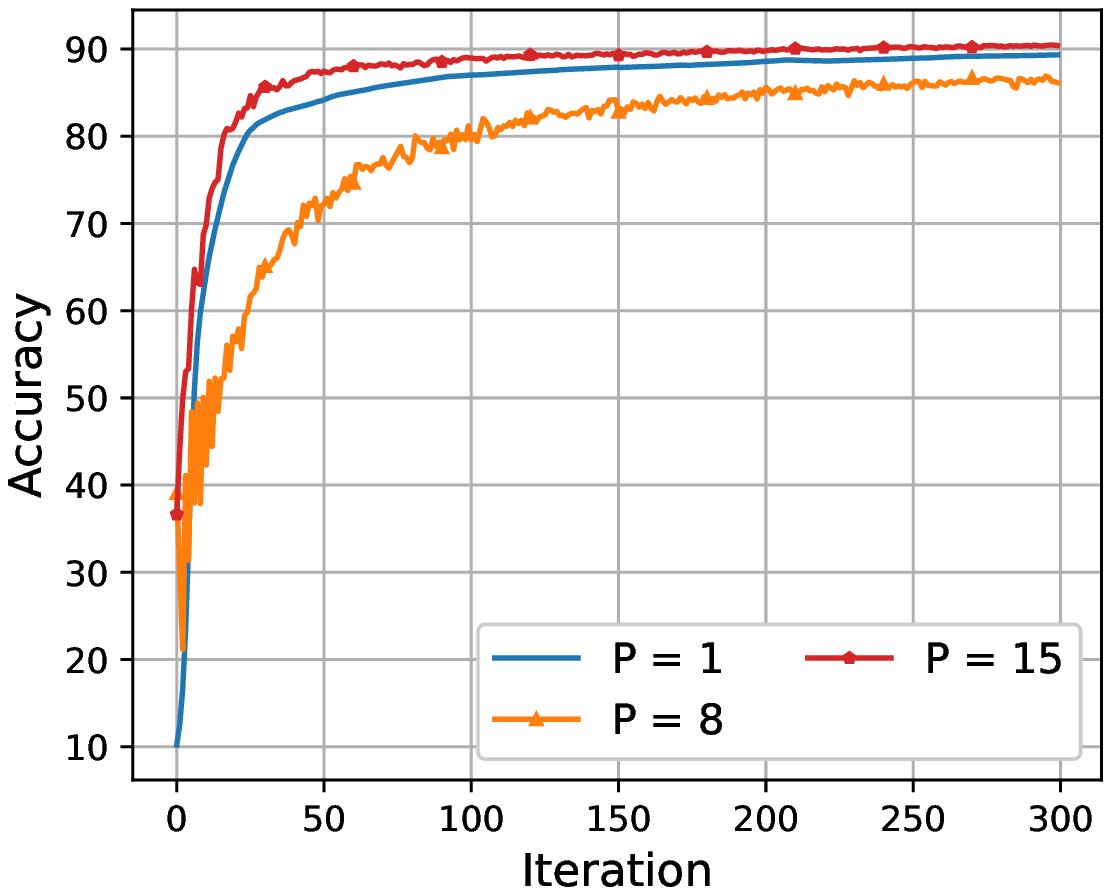}
    	\centerline{(c) Dirichlet=10}
    \end{minipage}
    \caption{Accuracy with Different Non-IID Degree}
    \label{fig:degree_acc}
\end{figure}
\begin{figure}[t]
    \centering
    \begin{minipage}[c]{0.15\textwidth}
    	\centering
    	\includegraphics[width=1.2in]{figures/density/Density-0.2-LOSS.eps}
    	\centerline{(a) Dirichlet=0.1}
    \end{minipage}
    \begin{minipage}[c]{0.15\textwidth}
    	\centering
    	\includegraphics[width=1.2in]{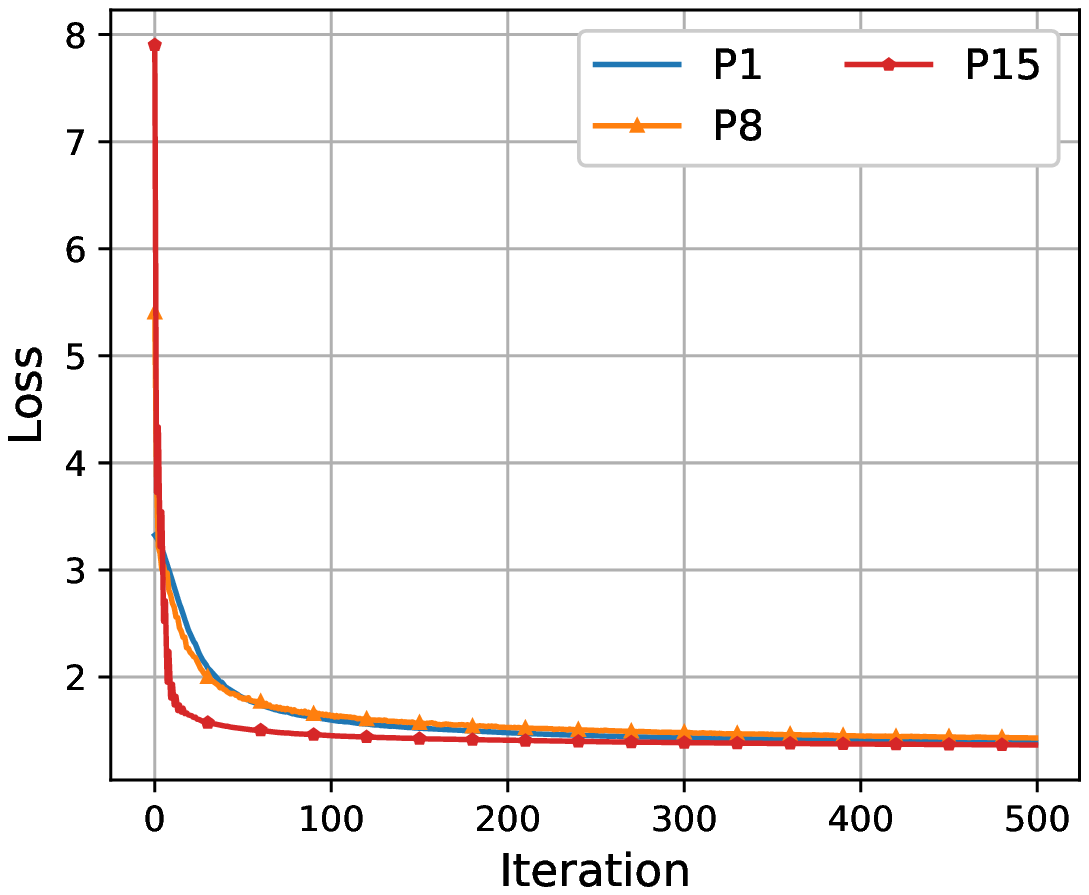}
    	\centerline{(b) Dirichlet=1}
    \end{minipage}
    \begin{minipage}[c]{0.15\textwidth}
    	\centering
    	\includegraphics[width=1.2in]{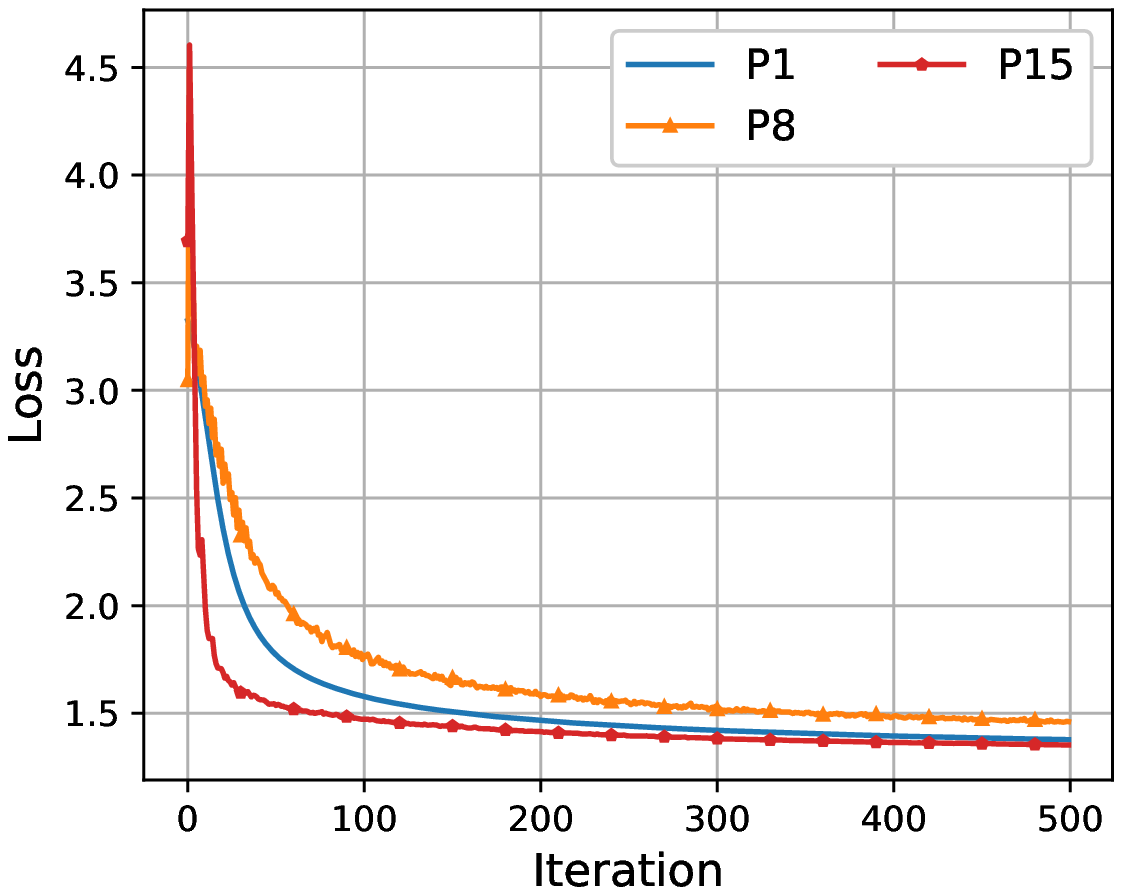}
    	\centerline{(c) Dirichlet=10}
    \end{minipage}
    \caption{Loss with Different Non-IID Degree}
    \label{fig:degree_loss}
\end{figure}
\begin{figure}[t!b]
    \centering
    \begin{minipage}[c]{0.15\textwidth}
    	\centering
    	\includegraphics[width=1.2in]{figures/density/Density-0.2-CS.eps}
    	\centerline{(a) Dirichlet=0.1}
    \end{minipage}
    \begin{minipage}[c]{0.15\textwidth}
    	\centering
    	\includegraphics[width=1.2in]{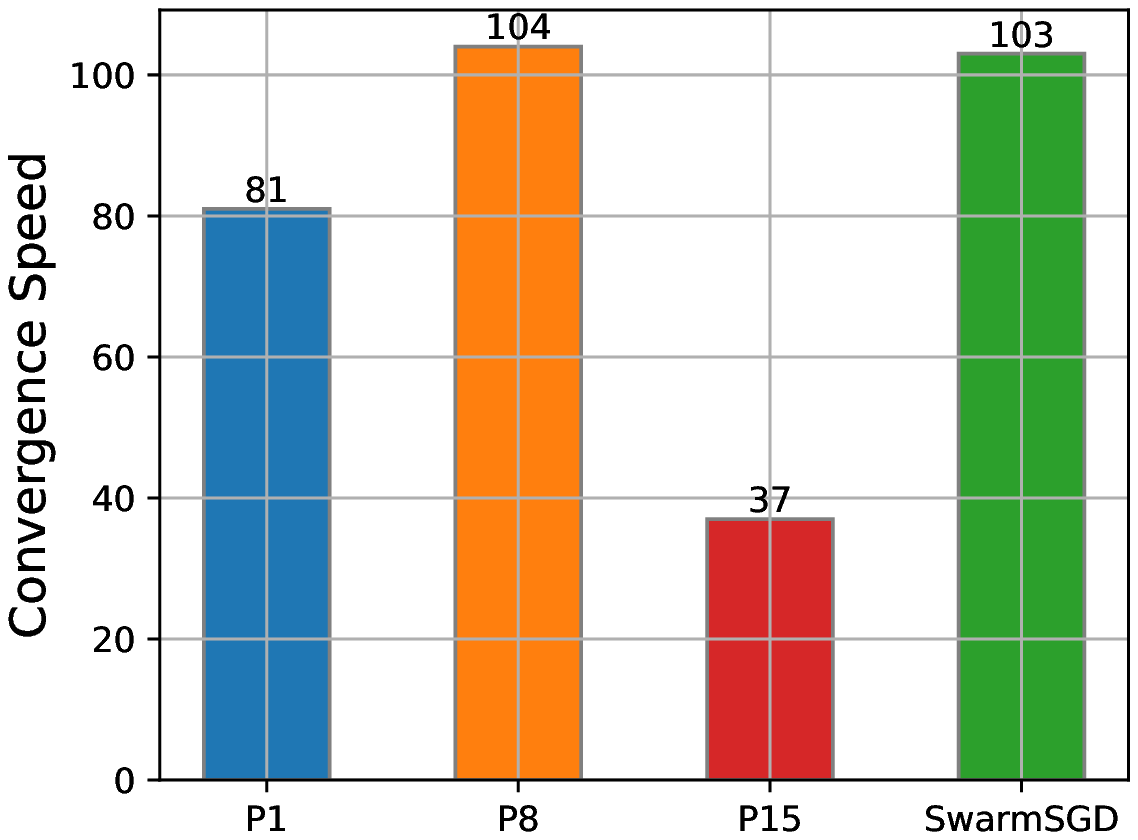}
    	\centerline{(b) Dirichlet=1}
    \end{minipage}
    \begin{minipage}[c]{0.15\textwidth}
    	\centering
    	\includegraphics[width=1.2in]{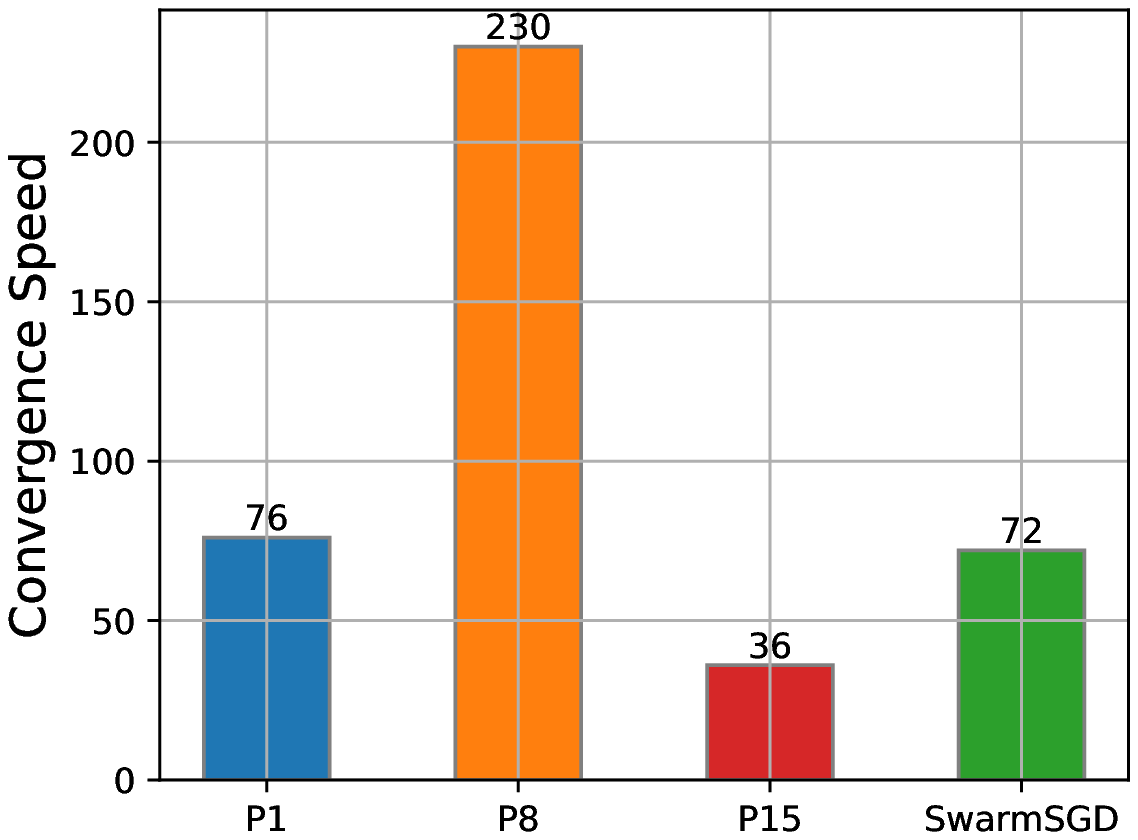}
    	\centerline{(c) Dirichlet=10}
    \end{minipage}
    \caption{Iterations to Converge with Different Non-IID Degree}
    \label{fig:degree_converge}
\end{figure}

\para{Convergence with Different Non-IID Degree.}
\label{subsubsec:different_P_comparison}
Third, we explore the relation between the performance of \sys{} and Non-IID Dirichlet Degree. We set $\alpha=\{0.1, 1, 10\}$, respectively. Results are shown in Figures \ref{fig:degree_acc}, \ref{fig:degree_loss}, \ref{fig:degree_converge}.

The results show that \sys{} significantly improved the convergence speed in Non-IID scenarios, especially in the extreme Non-IID setting where $\alpha=0.1$, \sys{} roughly increased the convergence speed by 40\%. The lower the $\alpha$, the better the performance of \sys{}.

\begin{figure}[t]
    \centering
    \begin{minipage}[c]{0.15\textwidth}
    	\centering
    	\includegraphics[width=1.2in]{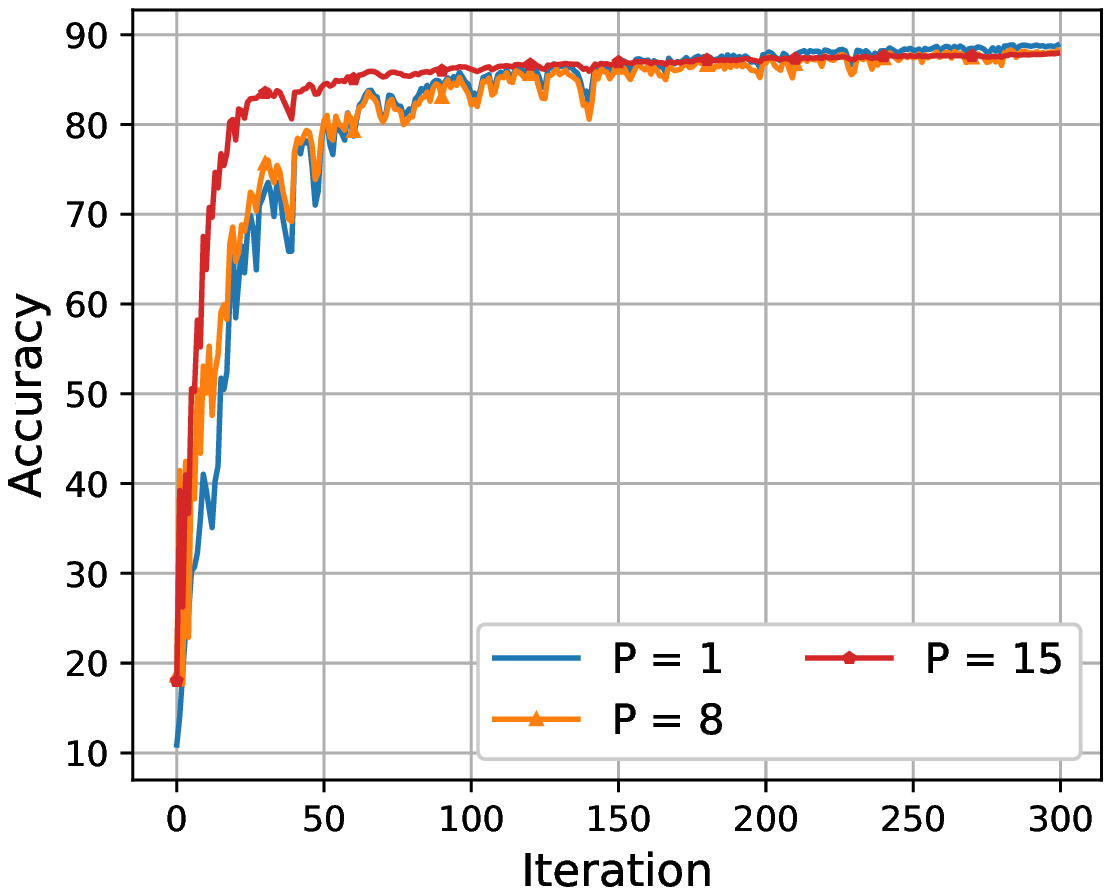}
    	\centerline{(a) $m$=5}
    \end{minipage}
    \begin{minipage}[c]{0.15\textwidth}
    	\centering
    	\includegraphics[width=1.2in]{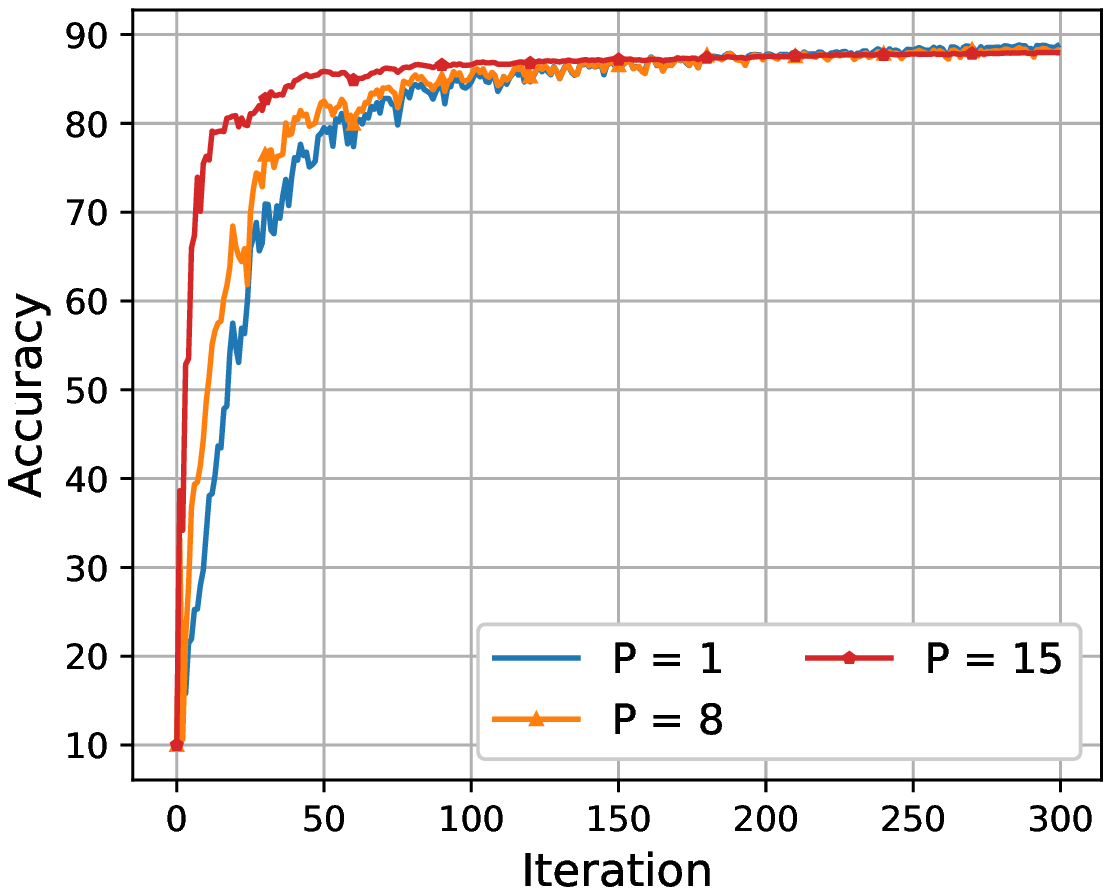}
    	\centerline{(b) $m$=10}
    \end{minipage}
    \begin{minipage}[c]{0.15\textwidth}
    	\centering
    	\includegraphics[width=1.2in]{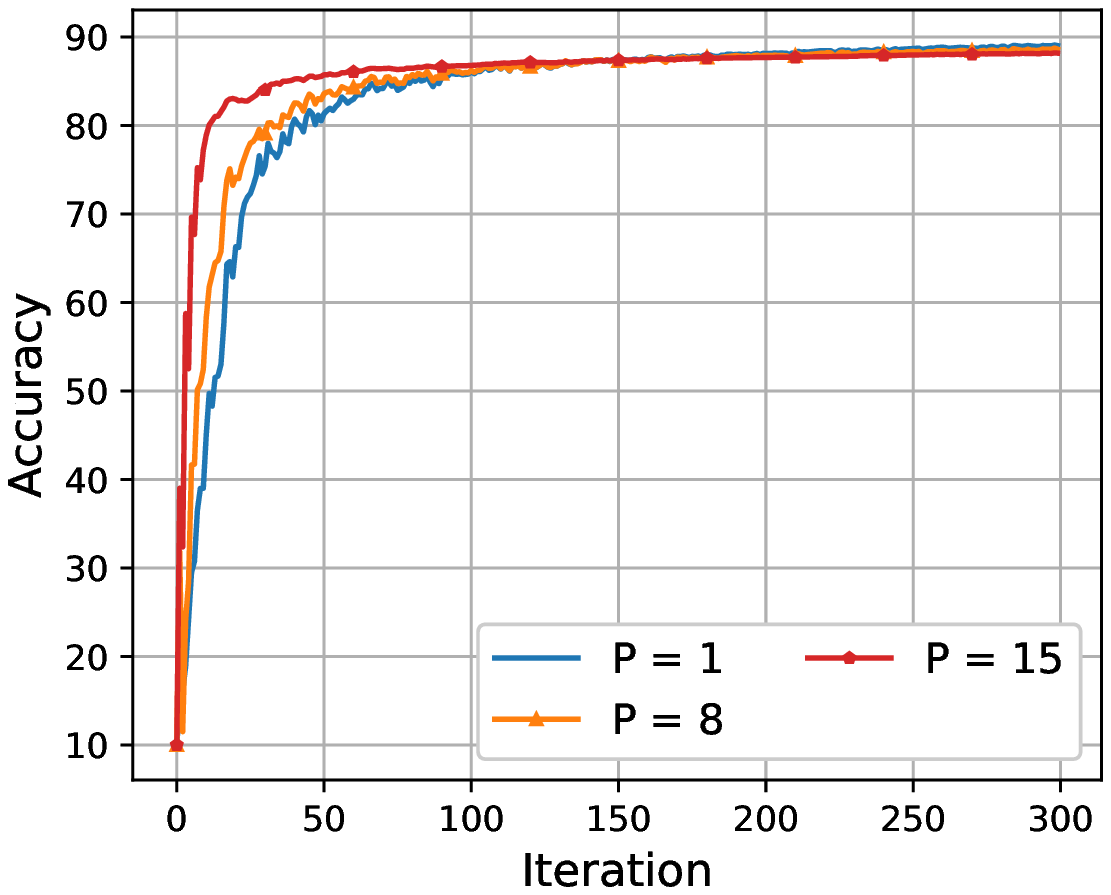}
    	\centerline{(c) $m$=20}
    \end{minipage}
    \caption{Accuracy with Different Environment Scale}
    \label{fig:scale_acc}
\end{figure}

\begin{figure}[t]
    \centering
    \begin{minipage}[c]{0.15\textwidth}
    	\centering
    	\includegraphics[width=1.2in]{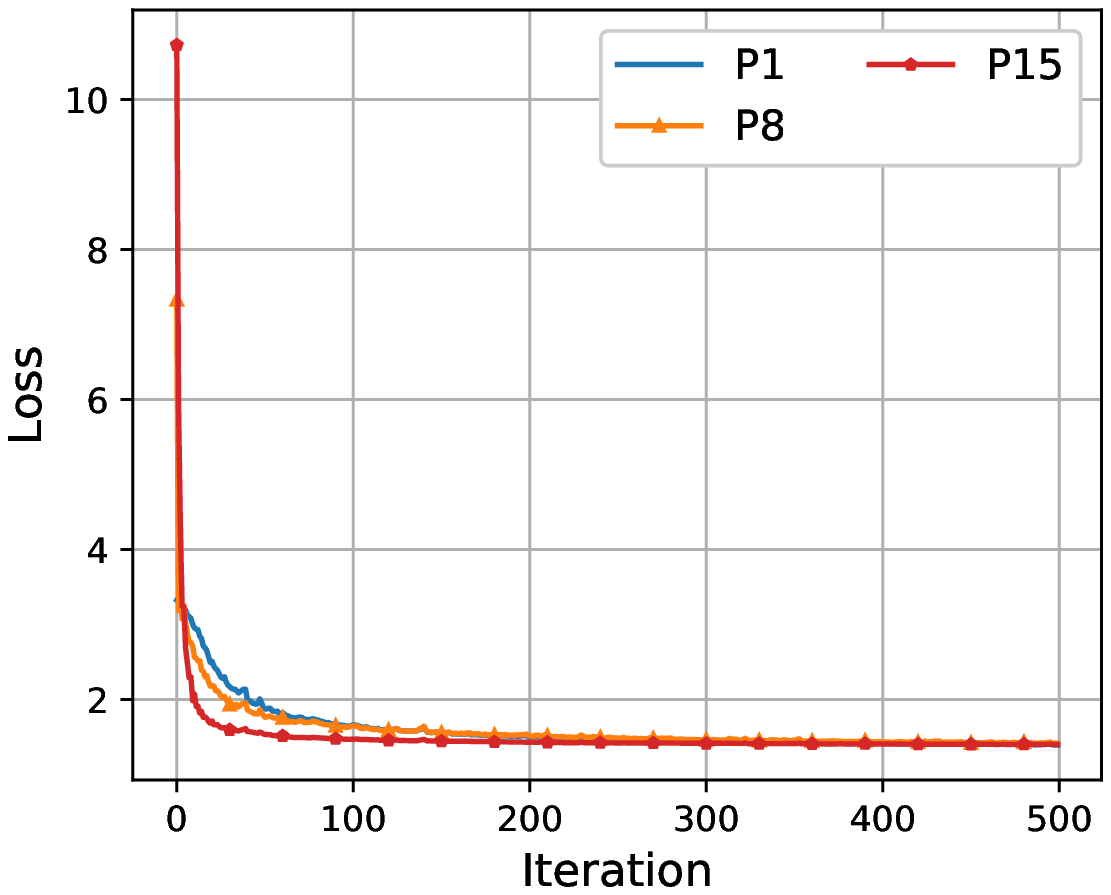}
    	\centerline{(a) $m$=5}
    \end{minipage}
    \begin{minipage}[c]{0.15\textwidth}
    	\centering
    	\includegraphics[width=1.2in]{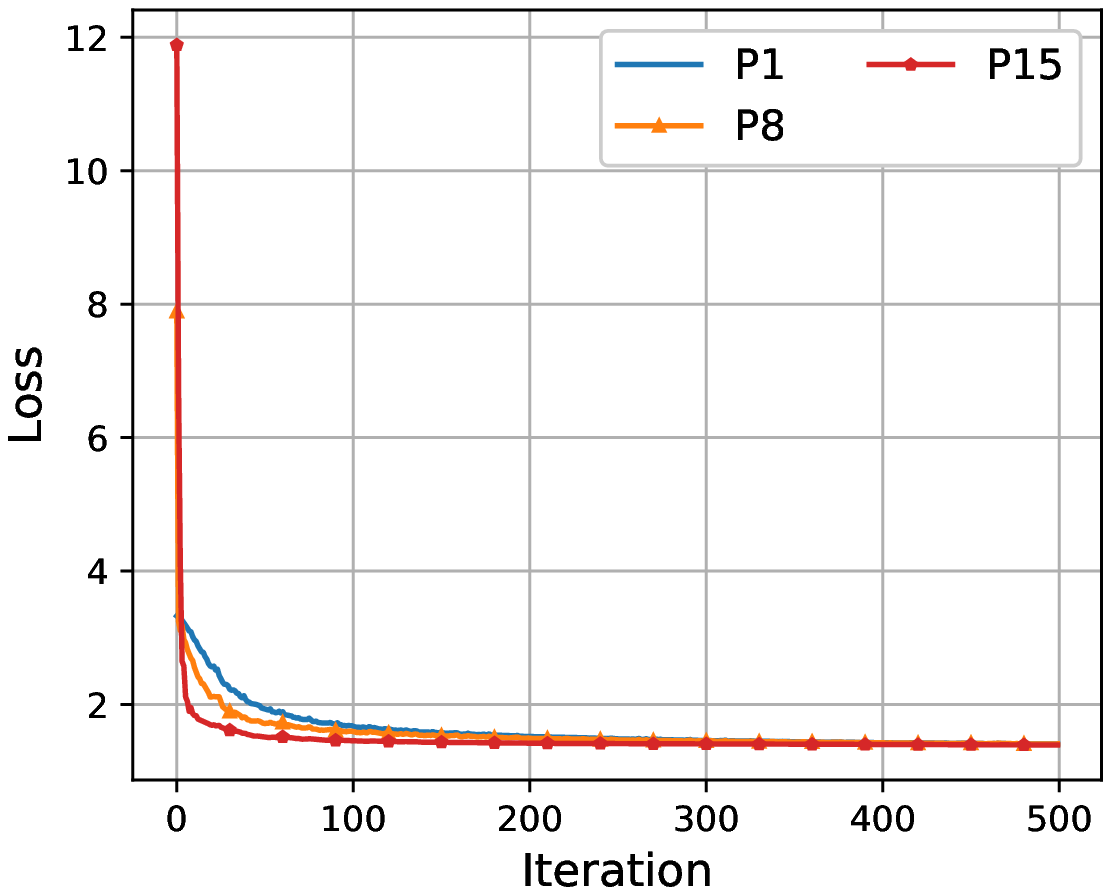}
    	\centerline{(b) $m$=10}
    \end{minipage}
    \begin{minipage}[c]{0.15\textwidth}
    	\centering
    	\includegraphics[width=1.2in]{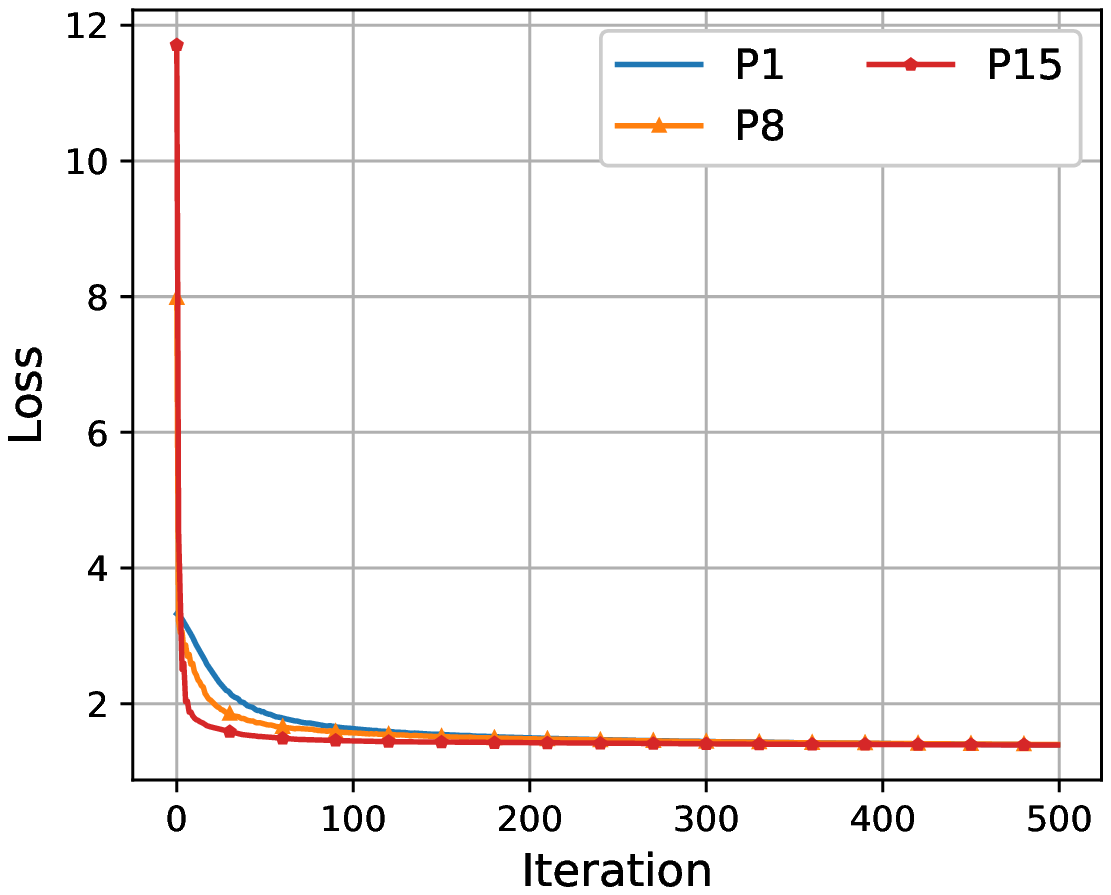}
    	\centerline{(c) $m$=20}
    \end{minipage}
    \caption{Loss with Different Environment Scale}
    \label{fig:scale_loss}
\end{figure}

\begin{figure}[t!b]
    \centering
    \begin{minipage}[c]{0.15\textwidth}
    	\centering
    	\includegraphics[width=1.2in]{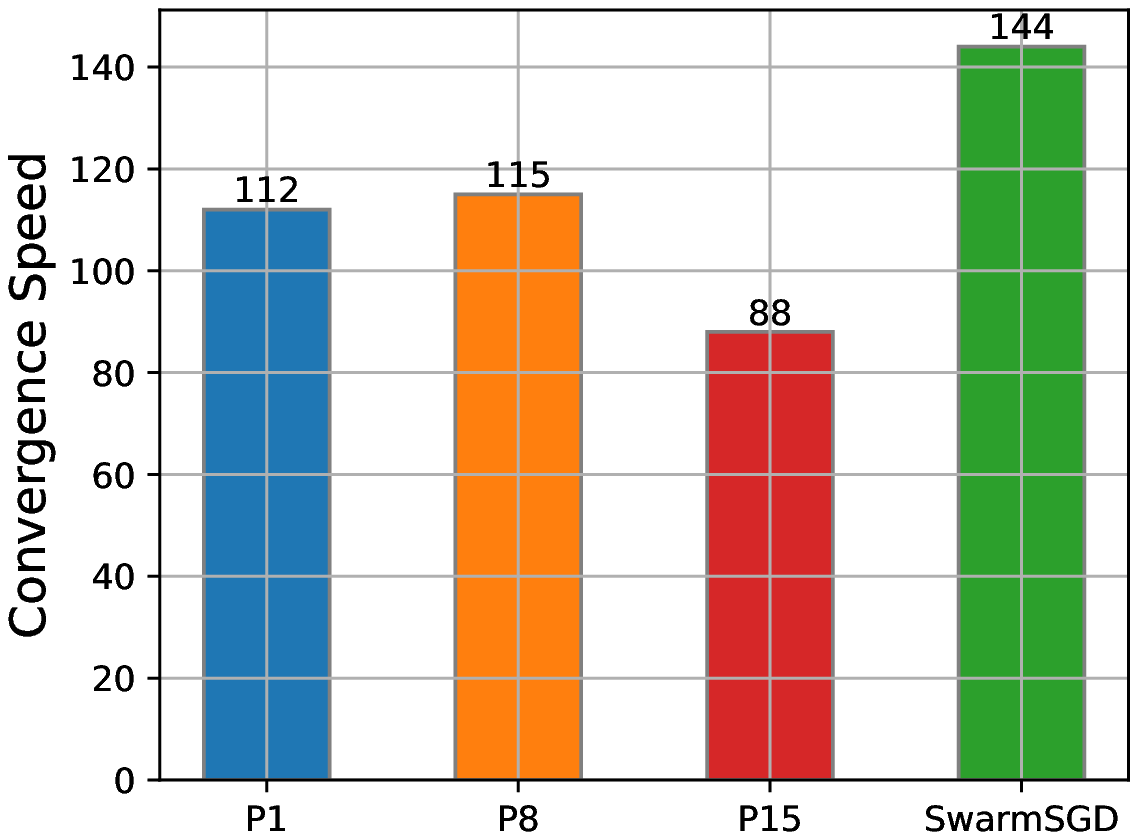}
    	\centerline{(a) $m$=5}
    \end{minipage}
    \begin{minipage}[c]{0.15\textwidth}
    	\centering
    	\includegraphics[width=1.2in]{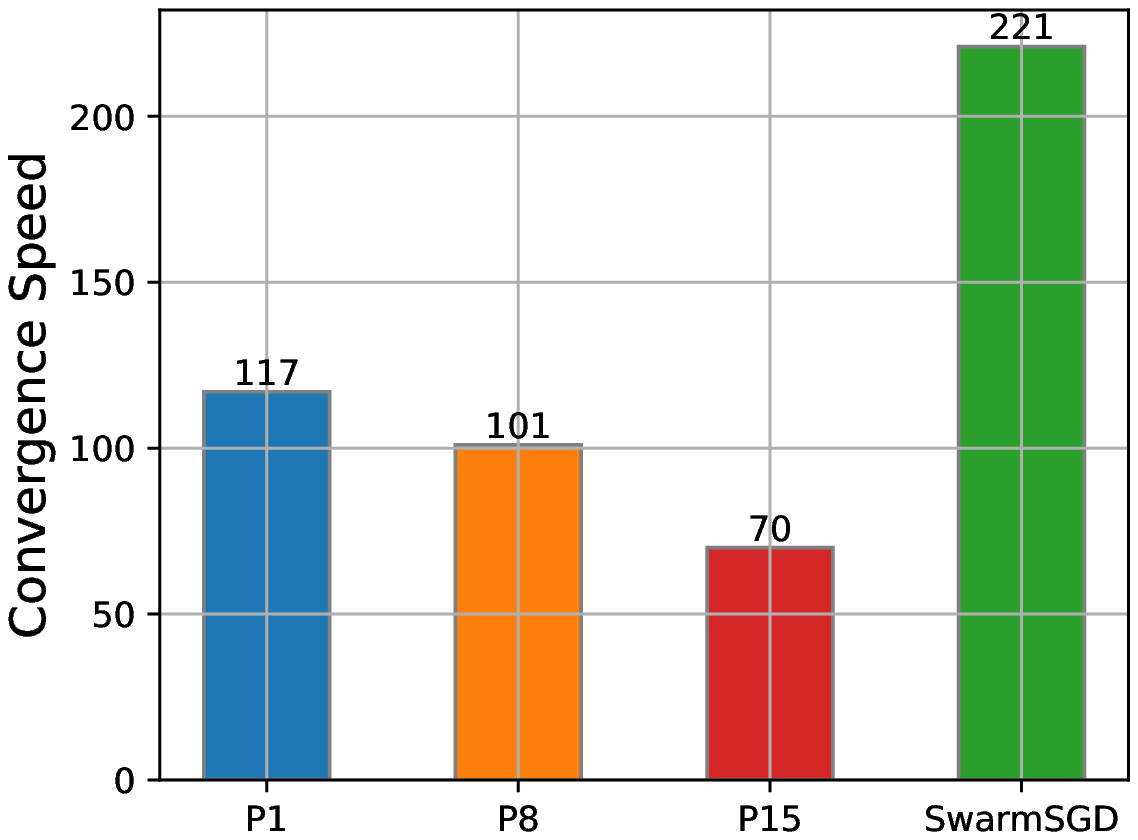}
    	\centerline{(b) $m$=10}
    \end{minipage}
    \begin{minipage}[c]{0.15\textwidth}
    	\centering
    	\includegraphics[width=1.2in]{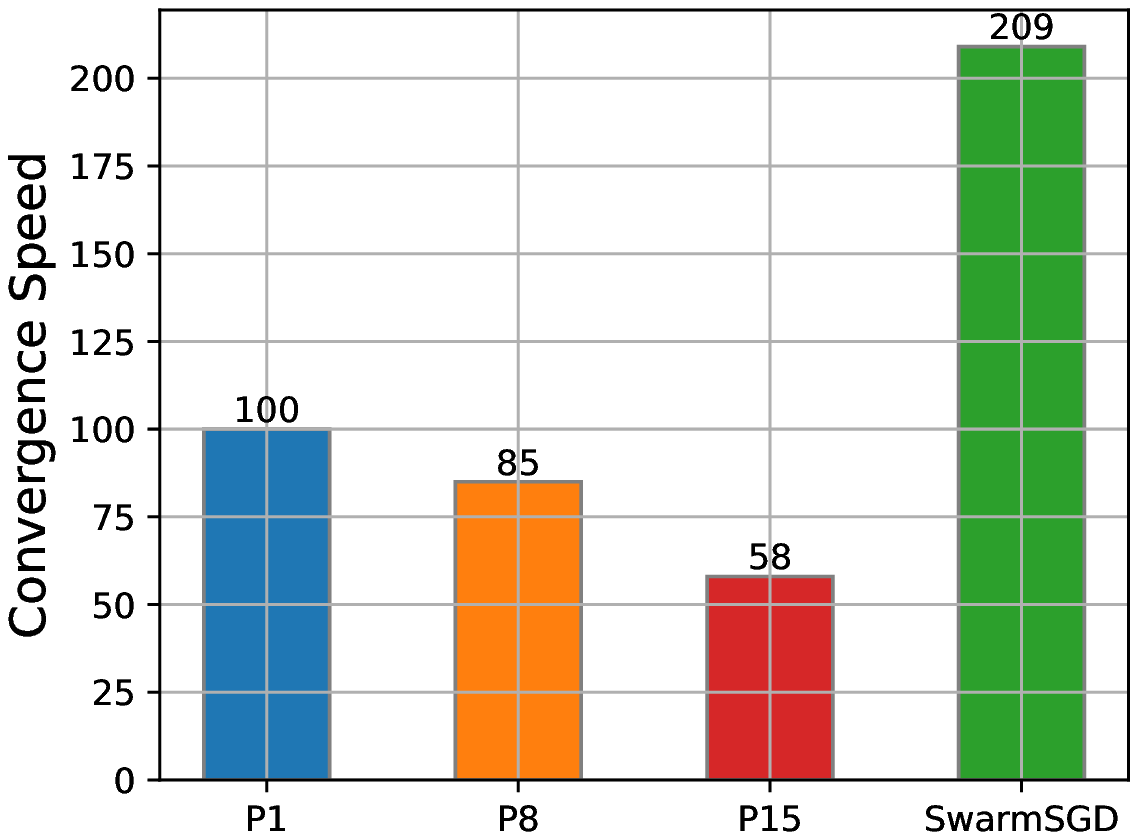}
    	\centerline{(c) $m$=20}
    \end{minipage}
    \caption{Converge Iterations with Different $m$ Values}
    \label{fig:scale_converge}
\end{figure}

\para{Convergence with Different Scalability.}
\label{subsubsec:different_scale}
We set different values of $m$ (numbers of devices) to find the impact of $m$ on performance. Specifically we set the number of devices as $m=\{5, 10, 20\}$. We plot the results in Figures \ref{fig:scale_acc}, \ref{fig:scale_loss}, \ref{fig:scale_converge}.

As shown \sys{} is also robustly scalable in weak-connectivity  environments. 
\sys{} still holds the best convergence speed in different $m$ values. In the meantime, other baselines such as SwarmSGD suffer heavily from the increasing $m$ due to its fewer aggregation rounds.
\section{Conclusion}
\label{sec:conclusion}
We studied the important problem of DML under \textit{divergence forces}: heterogeneous data and weak communication. To solve these, we introduce the novel \sys{} framework, where averaging occurs in a mirror space. We rigorously show a $O(m \sqrt[r-1]{\eta m} +\frac{m}{\eta T})$ bound on loss, which is $O(\sqrt[r]{\frac{m^{r+1}}{T}})$ with optimal $\eta$. For weak-connectivity situations (\ie, $m\geq T)$, selecting a weighted power mean of degree $p$ gives $r=p+1$, which can decrease loss by up to $O(\sqrt{m/T})$. Additionally, through experimental validation, we show that \sys{} improves DML's convergence speed in all cases, and improves model accuracy under weak-communication, Non-IID data, or large numbers of devices.

\bibliography{ref}
\bibliographystyle{IEEEtran}

\clearpage
\section{Appendix}
\label{sec:appendix}

\subsection{Proof of Lemma \ref{lemma:consensus}}
\label{subsec:lemma1}
We are able to write out a general formula for $\nabla \phi (\textbf{w}_{i,t+1})$:
\begin{equation}
\label{eq:17}
\begin{aligned}
 &\nabla \phi(\textbf{w}_{i,t+1}) = \sum_{j=1}^m P(t,k)_{ij} \nabla \phi(\textbf{w}_{j,k})\\
 &-\eta \sum_{\tau=k+1}^t \sum_{j=1}^m P(t,\tau)_{ij} \cdot \nabla f_{j}(\textbf{w}_{j,\tau-1}) - \eta \nabla f_i(\textbf{w}_{i,t}).
\end{aligned}
\end{equation}
as well as $\nabla \phi(\overline{\textbf{w}_t})$:
\begin{equation}
\label{eq:18}
\begin{aligned}
&\nabla \phi(\overline{\textbf{w}_t}) = \frac1m \sum_{i=1}^m \nabla \phi (\textbf{w}_{i,t}) = \sum_{j=1}^m \frac1m \nabla \phi(\textbf{w}_{j,k})\\
&-\eta \sum_{\tau=k+1}^t \sum_{j=1}^m \frac1m \cdot \nabla f_{j}(\textbf{w}_{j,\tau-1}) - \frac{\eta}{m}\sum_{i=1}^m \nabla f_i(\textbf{w}_{i,t}).
\end{aligned}
\end{equation}

Then, $\nabla \phi(\textbf{w}_{i,t}) - \nabla \phi(\overline{\textbf{w}_t})$ can be bounded by applying the Triangle Inequality and Equation (\ref{eq:expmix}):
\begin{equation}
\begin{aligned}
    &\lVert \nabla \phi(\textbf{w}_{i,t+1}) - \nabla \phi(\overline{\textbf{w}_{t+1}})\rVert\leq  \sum_{j=1}^m \vartheta \kappa^{t-k}\lVert \nabla \phi(\textbf{w}_{j,k})\rVert\\
    &+\sum_{\tau=k+1}^t \sum_{j=1}^m \vartheta \kappa^{t-\tau} \lVert \eta \nabla f_{j}(\textbf{w}_{j,\tau-1})\rVert + 2\eta G_l.\\
\end{aligned}
\end{equation} 
Then, plugging in $k=0$ gives
\begin{equation}
\begin{aligned}
    &\leq  \vartheta (\kappa^{t} \sum_{j=1}^m \lVert \nabla \phi(\textbf{w}_{j,0})\rVert + m\eta G_l \sum_{\tau=1}^t \kappa^{t-\tau} + 2\eta G_l)\\
    &\leq \vartheta( \kappa^{t} \sum_{j=1}^m \lVert \nabla \phi(\textbf{w}_{j,0})\rVert + m\eta G_l \cdot \frac{1}{1-\kappa} + 2\eta G_l).
\end{aligned}
\end{equation}
Finally, shifting $t$ down by 1 gives the desired bound.

\subsection{Proof of Theorem \ref{thm:maintheorem}}
\label{subsec:theorem1}
We first prove a claim:
\begin{claim}
Under Algorithm \ref{alg:new} and Assumption \ref{ass:boundedgradient}:
\label{claim:wamgm}

  \begin{align*}
    &(m\eta G_l) \cdot \lVert \overline{\textbf{w}_t} - \overline{\textbf{w}_{t+1}}\rVert\\
    &\leq (r-1)\cdot \sqrt[r-1]{\frac{1}{\sigma  r^{r-1}m}\cdot (m\eta G_l)^r} + m\cdot \frac{\sigma}{r} \lVert \overline{\textbf{w}_t} - \overline{\textbf{w}_{t+1}}\rVert^r.
\end{align*}
\end{claim}
\begin{proof}
Weighted AM-GM gives:
\begin{align*}
    &(r-1)\cdot \sqrt[r-1]{\frac{1}{\sigma  r^{r-1}m}\cdot (m\eta G_l)^r} + m\cdot \frac{\sigma}{r} \lVert \overline{y_t} - \overline{y_{t+1}}\rVert^r\\
    &\geq r\cdot \left(\sqrt[r-1]{\frac{1}{\sigma  r^{r-1}m}\cdot (m\eta G_l)^r})^{r-1}\cdot (m\cdot \frac{\sigma}{r} \lVert \overline{y_t} - \overline{y_{t+1}}\rVert^r)^1\right)^{\frac1r}\\
    &=r\cdot \left(\frac{1}{ r^{r}}\cdot (m\eta G_l)^r\cdot  \lVert \overline{y_t} - \overline{y_{t+1}}\rVert^r\right)^{\frac1r} = (m\eta G_l)\cdot \lVert \overline{y_t} - \overline{y_{t+1}}\rVert.
\end{align*}

\end{proof}
We may follow the main line of reasoning that proves Theorem \ref{thm:maintheorem}.
\subsection{Main Line of Reasoning}
\begin{proof}
We prove bounds for generic $x$. Note that $\overline{\textbf{w}_{t}} = h^{-1} (\frac1m \sum_{i=1}^m h(\textbf{w}_{i,t}))$. Then, we get:
\begin{equation}
\begin{aligned}
\label{eq:firstbound}
    &\eta \sum_{i=1}^m f_{i}(\textbf{w}_{i,t}) - f_{i}(x) \leq \sum_{i=1}^m \langle \eta \nabla f_i (\textbf{w}_{i,t}), \textbf{w}_{i,t}-x\rangle\\
    &=  \sum_{i=1}^m \langle \eta \nabla f_i(\textbf{w}_{i,t}) , \textbf{w}_{i,t} - \overline{\textbf{w}_{t}}\rangle \\ 
    &+\sum_{i=1}^m \langle \eta \nabla f_i(\textbf{w}_{i,t}) , \overline{\textbf{w}_{t}} - \overline{\textbf{w}_{t+1}}\rangle  +\langle \eta \nabla f_i(\textbf{w}_{i,t}) , \overline{\textbf{w}_{t+1}}-x\rangle.
\end{aligned}    
\end{equation}

Cauchy's inequality can bound the first term. The third term can be manipulated using Equation (\ref{eq:gradphi}). Then, combined with $f_i(\overline{\textbf{w}_t}) \leq f_i(\textbf{w}_{i,t}) + G_l \lVert \overline{\textbf{w}_t}-\textbf{w}_{i,t}\rVert$, we get
\begin{align*}
&\eta\cdot (F(\overline{\textbf{w}_t}) - F(x)) \leq (2\cdot \sum_{i=1}^m \eta G_l \cdot \lVert \textbf{w}_{i,t} - \overline{\textbf{w}_{t}}\rVert) \\
&+ m\eta G_l \cdot \lVert \overline{\textbf{w}_t}-\overline{\textbf{w}_{t+1}}\rVert + \sum_{i=1}^m\langle \nabla \phi(y_{i,t}) - \phi(\textbf{w}_{i,t+1}) , \overline{\textbf{w}_{t+1}} - x\rangle.
\end{align*}
The factor of 2 comes from the 2nd term of Eq (\ref{eq:firstbound}) added to the error from $f_i(\overline{\textbf{w}_t}) \leq f_i(\textbf{w}_{i,t}) + G_l \lVert \overline{\textbf{w}_t}-\textbf{w}_{i,t}\rVert$. Note that the last term is equal to $m\cdot \langle \overline{\textbf{w}_t}-\overline{\textbf{w}_{t+1}},\overline{\textbf{w}_{t+1}}-x\rangle$. This is also equal to $m(D_{\phi}(x,\overline{\textbf{w}_t}) - D_{\phi}(x,\overline{\textbf{w}_{t+1}}) - D_{\phi}(\overline{\textbf{w}_{t+1}}, \overline{\textbf{w}_t}))$ by the Triangle Inequality for Bregman Divergences. We also substitute Claim 1 to replace the second term, so the value is
\begin{equation}
  \begin{aligned}
    &\leq (2 \cdot \sum_{i=1}^m \eta G_l \cdot \lVert \textbf{w}_{i,t} - \overline{\textbf{w}_t}\rVert) \\
    &+\frac{r-1}{r}\sqrt[r-1]{\frac{1}{\sigma m}\cdot (m\eta G_l)^r}  + \sigma \lVert \overline{\textbf{w}_t} - \overline{\textbf{w}_{t+1}}\rVert^r\\
    &+ m(D_{\phi}(x,\overline{\textbf{w}_t}) - D_{\phi}(x,\overline{\textbf{w}_{t+1}}) - D_{\phi}(\overline{\textbf{w}_{t+1}}, \overline{\textbf{w}_t})).
\end{aligned}  
\end{equation}

But, by uniform convexity, $D_{\phi}(\overline{\textbf{w}_{t+1}}, \overline{\textbf{w}_t}) \geq \sigma \lVert \overline{\textbf{w}_{t+1}}- \overline{\textbf{w}_t}\rVert^r$, and thus this is also
\begin{equation}
  \begin{aligned}
    &\leq (2\cdot \sum_{i=1}^m \eta G_l \cdot \lVert \textbf{w}_{i,t} - \overline{\textbf{w}_{t}}\rVert)+\frac{r-1}{r}\sqrt[r-1]{\frac{1}{\sigma m}\cdot (m\eta G_l)^r}\\
    &+ m(D_{\phi}(x,\overline{\textbf{w}_t}) - D_{\phi}(x,\overline{\textbf{w}_{t+1}})).
\end{aligned}  
\end{equation}
Then, taking the sum over $T$ gives
\begin{equation}
  \begin{aligned}
&\eta \sum_{t=0}^{T-1} F(\overline{\textbf{w}_t}) - F(x) \leq \sum_{t=0}^{T-1} 2\cdot( \sum_{i=1}^m \eta G_l \cdot \lVert \textbf{w}_{i,t} - \overline{\textbf{w}_{t}}\rVert)\\
&+T\cdot \frac{r-1}{r}\sqrt[r-1]{\frac{1}{\sigma m}\cdot (m\eta G_l)^r}\\
&+ m(D_{\phi}(x,\overline{\textbf{w}_0}) - D_{\phi}(x,\overline{\textbf{w}_{T}}) )\\
&\leq 2m\eta G_l \cdot \sum_{t=0}^{T-1} \sqrt[r-1]{\frac{1}{\sigma} \lVert \nabla \phi(\textbf{w}_{i,t}) - \nabla \phi(\overline{\textbf{w}_t})\rVert }\\
&+Tm\cdot \frac{r-1}{r}  \sqrt[r-1]{\frac{1}{\sigma }\cdot (\eta G_l)^r}+ m\cdot D_{\phi}(x,\overline{\textbf{w}_0}).
\end{aligned}
\end{equation}
Dividing through by $\eta T$, substituting $x=x^{\ast}$, and substituting Lemma \ref{lemma:consensus} gives the desired result.
\end{proof}

\subsection{Skew Correction}
\begin{theorem}
Consider positive $x_i$ and $\alpha_i\approx \frac1m $. Then,
\begin{equation}
\left(\sum_{i=1}^m \alpha_i x_i\right)^{\frac1p} \leq \left(\sum_{i=1}^m \frac1m x_i\right)^{\frac1p} \cdot \left(1+ \frac{m\cdot \max_i |\alpha_i-\frac1m|}{p}\right)
\end{equation}
This allows us to see the weighted power mean as a way to decrease skew; this theorem is relevant in the setting of Lemma \ref{lemma:consensus}.
\end{theorem}


\begin{proof}
Write $\alpha_i = \frac1m + \epsilon \cdot e_i$ where where $\epsilon = m\cdot \max_i |\alpha_i-\frac1m|$. Then, for all $i$, we have $|e_i|\leq \frac1m$. 

As $\alpha\to \frac1m$, we also have $\epsilon \to 0$. The derivative of the weighted power mean with respect to $\epsilon$ is 
\[\frac{\sum e_i x_i^p }{p\cdot\left(\sum \frac1m x_i^p +  \epsilon\cdot(\sum e_i x_i^p)\right)^{\frac{p-1}{p}}}\]
and the derivative at $\epsilon=0$ is $\frac{\sum e_i x_i^p }{p\cdot\left(\sum \frac1m x_i^p\right)^{\frac{p-1}{p}}}$. Thus, we have the following first order approximation when $\epsilon \approx 0$ due to the concavity with respect to $\epsilon$:
\begin{align}
    \left(\sum_{i=1}^m \alpha_i x_i\right)^{\frac1p} &\leq \left(\sum_{i=1}^m \frac1m x_i\right)^{\frac1p} + \epsilon \cdot \frac{\sum e_i x_i^p }{p\cdot\left(\sum \frac1m x_i^p\right)^{\frac{p-1}{p}}}
\end{align}
But, since $x_i\geq 0$ and $e_i\leq \frac1m$, we have $\sum e_i x_i^p \leq \sum \frac1m x_i^p$, so 
\begin{align}
    \left(\sum_{i=1}^m \alpha_i x_i\right)^{\frac1p} &\leq \left(\sum_{i=1}^m \frac1m x_i\right)^{\frac1p} + \epsilon \cdot \frac{1}{p} \left(\sum \frac1m x_i^p\right)^{\frac1p}
\end{align}
Thus, the main dependency is a skew on the order of $\frac{\epsilon}{p}$ where $\epsilon = m\cdot \max_i |\alpha_i-\frac1m|$.
\end{proof}

The main trouble with Gossip is that the $\left(\sum_{i=1}^m \alpha_i x_i\right)^{\frac1p}$ terms differ from the global average $\left(\sum_{i=1}^m \frac1m x_i\right)^{\frac1p}$. This Theorem shows how \sys{}, under $h(x)=x^p$, can reduce skew by a factor of $p$.

\end{document}